\documentclass{article}
\usepackage{physics}
\usepackage[utf8]{inputenc} % allow utf-8 input
\usepackage[T1]{fontenc}    % use 8-bit T1 fonts
\usepackage{hyperref}       % hyperlinks
\usepackage{url}            % simple URL typesetting
\usepackage{booktabs}       % professional-quality tables
\usepackage{amsfonts}       % blackboard math symbols
\usepackage{nicefrac}       % compact symbols for 1/2, etc.
\usepackage{microtype}      % microtypography
\usepackage{xcolor}         % colors
\usepackage[ruled]{algorithm2e}
\usepackage{amsmath}
\usepackage[pdftex]{graphicx}
\usepackage{colortbl} % For coloring cells
\usepackage{xcolor} % Define custom colors
\usepackage{array}
\newtheorem{definition}{Definition}
\usepackage{epstopdf}
\usepackage{authblk}

\title{Physics Informed Constrained Learning of Dynamics from Static Data}

\author[1]{Pengtao Dang\thanks{dangpe@ohsu.edu}}
\author[1]{Tingbo Guo}
\author[2]{Melissa Fishel}
\author[3]{Guang Lin}
\author[3]{Wenzhuo Wu}
\author[1]{Sha Cao}
\author[1]{Chi Zhang\thanks{zhangchi@ohsu.edu}}

\affil[1]{Oregon Health $\And$ Science University}
\affil[2]{Indiana University}
\affil[3]{Purdue University}

\date{}

\begin{document}

\maketitle

\begin{abstract}
A physics-informed neural network (PINN) models the dynamics of a system by integrating the governing physical laws into the architecture of a neural network. By enforcing physical laws as constraints, PINN overcomes challenges with data scarsity and potentially high dimensionality. Existing PINN frameworks rely on fully observed time-course data, the acquisition of which could be prohibitive for many systems. In this study, we developed a new PINN learning paradigm, namely Constrained Learning, that enables the approximation of first-order derivatives or motions using non-time course or partially observed data. Computational principles and a general mathematical formulation of Constrained Learning were developed. We further introduced MPOCtrL (Message Passing Optimization-based Constrained Learning) an optimization approach tailored for the Constrained Learning framework that strives to balance the fitting of physical models and observed data. Its code is available at github link: \url{https://github.com/ptdang1001/MPOCtrL}. Experiments on synthetic and real-world data demonstrated that MPOCtrL can effectively detect the nonlinear dependency between observed data and the underlying physical properties of the system. In particular, on the task of metabolic flux analysis, MPOCtrL outperforms all existing data-driven flux estimators.
\end{abstract}

\section{Introduction}
\label{introduction}

Physics-informed neural networks (PINNs) are a class of machine learning models that integrate physical laws into the training process of neural networks \cite{raissi2019physics,dang2023generalized,karniadakis2021physics}. Unlike traditional neural networks that rely solely on data for training, PINNs incorporate known physical principles to guide the learning process, which has been shown to improve the model's accuracy and generalizability, especially in scenarios where data is scarce, noisy, or potentially high-dimensional
\cite{raissi2019physics,karniadakis2021physics}. In the realm of biological sciences, PINNs gained its popularity given the fundamental challenge to discover the mathematical equations that govern  biological systems from observed data \cite{yazdani2020systems,daneker2023systems,daryakenari2024ai}. From neuroscience to biomechanics, from pharmacokinetics to metabolic flux analysis, PINNs offer a versatile framework for capturing intricate interactions and dynamics within biological systems \cite{sarabian2022physics,daryakenari2024ai,zhang2022physics,orth2010flux}. For instance, PINNs have been utilized to model neuronal activity and network dynamics, to shed light on the underlying mechanisms of cognition and behavior \cite{jiang2022physics,wang2021physics}. Moreover, PINNs have facilitated the simulation of tissue biomechanics, to aid in the design of prosthetics and rehabilitation strategies \cite{zhang2022physics}. Additionally, the idea of PINNs has been instrumental in deciphering the kinetics of metabolic pathways, and offering insights into disease mechanisms \cite{orth2010flux}.

Despite its groundbreaking capability, PINNs require time course data as input to effectively capture temporal dynamics via differential equations. However, for many systems, only static or snapshot data is available, diminishing the efficacy of PINNs. Moreover, PINNs always rely on the measurement of all relevant variables in dynamic systems for accurate approximation, posing challenges in scenarios where obtaining such data is expensive, impractical, or limited. For instance, in the approximation of traffic flow \cite{pujawan2015integrated}, the dependence on extensive high-speed camera measures can impede the modeling process. Instead, low-speed or one-shot photos or even noise or dust measures on each specific road can be applied to approximate traffic flows. Similarly, in biological systems, such as human disease tissue metabolism, detailed temporal observations are often unattainable, and many variables may only be measured in a subset of study subjects. The challenges of working with partially observed, non-time course data underscore the need for novel approaches to extend the applicability of PINNs to a broader range of scientific and engineering problems.

In this study, we propose a new learning framework, namely Constrained Learning, to address the limitations of PINNs and enhance their applicability when time-course data and direct measurement of key variables in the system are unavailable. We begin by presenting the mathematical principles underpinning Constrained Learning and its general mathematical formulation. Recognizing that Constrained Learning does not fit within the traditional paradigms of supervised or unsupervised learning, we developed a new optimization algorithm, named MPO-SL, that can optimize general models of Constrained Learning. We further demonstrate the application of this framework to a specific problem: the estimation of mass-carrying flux over a network using non-time course and partially observed data. We introduce a new PINN architecture, named MPOCtrL, to approximate flux rates over a network by leveraging the governing physical laws of the system to the fitting of the observed data.  Experiments using synthetic and real-world data demonstrated that the MPOCtrL model and the MPO-SL algorithm can effectively estimate the flux over the networks of different topological properties using non-time course and partially observed data, thereby validating the feasibility of the Constrained Learning framework.

The key contributions of this study are summarized as follows and benchmarked with experiments:\\
  1. Development of a \textbf{new learning paradigm}, namely Constrained Learning, to approximate dynamic models using non-time course and partially observed data. \\ 
  2. Design of a \textbf{new optimization algorithm} to effectively and efficiently solve the general Constrained Learning problems. \\
  3. Development of a \textbf{new PINN architecture}, namely MPOCtrL, to estimate flux over complex networks using non-time course and partially observed data.

\section{Related Work}
\label{related_work}

\subsection{Physics Informed Neural Networks (PINNs)} 

%Overall, by integrating data-driven insights with physics-based constraints, PINNs offer a unique framework for solving inverse problems, finding wide-ranging applications in mechanical, material, environmental, and biological sciences \cite{Orth2010, Palsson2015}.

% Constrained learning aims to identify non-linear functions capable of approximating the dynamic properties of a system using partially observed data.
Initially proposed by Raissi et al. \cite{raissi2019physics}, PINNs offer a unique framework for solving inverse problems by integrating data-driven insights with physics-based constraints. It has gained popularity across various scientific fields such as fluid dynamics \cite{raissi2020hidden,sun2020surrogate,mathews2021uncovering} and material science \cite{fang2019deep,chen2020physics,zhang2022analyses}.  %This integration enhances the prediction accuracy by leveraging known physical laws within neural networks, alleviating the need for large datasets, which often bottleneck traditional machine learning applications. 
However, despite their potential, training PINNs can often be computationally intensive \cite{wang2020understanding}, and the performance of PINNs can significantly vary with the choice of hyperparameters and the network architecture, raising concerns on robustness \cite{cai2021physics}. Ongoing research continues to address these challenges, and tools like DeepXDE  \cite{lu2021deepxde} and hybrid models that combine traditional numerical solvers with neural networks \cite{meng2020composite} advanced the accessibility and efficiency of PINNs. %This hybrid approach not only mitigates some of the computational burdens but also improves the generalization of PINNs across varied applications.
Improving the scalability of PINNs to tackle multi-physics and multi-scale problems is also garnering increasing attention \cite{karniadakis2021physics}. 
Training PINNs typically requires the input of time-course data. However, obtaining the necessary time-course or spatial data for PINNs is challenging for many research areas, complicating their applications \cite{wagner2021metabolic,alghamdi2021graph}. The development of PINNs for datasets that lack time-series information presents a significant challenge, but also holds immense potential if successful. %This task is particularly complex because it requires the adaptation of PINNs to handle static data alone, without the temporal information that usually guide the learning of dynamic systems. %Researchers are focused on innovating ways to integrate physical laws into neural networks when only steady observations are available. 
%This advancement could significantly broaden the applicability of PINNs in fields like material science, and static biological reaction analysis, where time-dependent data may not be available or relevant. The progress in this area is keenly watched as it could lead to more robust models\cite{jeong2023physics}. %As the computational power continues to grow and methodologies evolve, PINNs are set to play a pivotal role in the future of computational science, blending data-driven insights with established physical or biological models.

\subsection{Data Driven Flux Analysis}\label{Data_Driven_Flux_Analysis}
The flux dynamics of a system are typically described using differential equations to quantify the rate at which mass, energy, or momentum passes through a surface or region. In chemical diffusion, this is governed by Fick's laws, as seen in reaction-diffusion systems \cite{ji2021stiff}. In fluid dynamics, the Navier-Stokes equations apply, such as in the modeling of blood flow to identify key hemodynamic biomarkers of cardiovascular diseases \cite{arzani2021uncovering,vardhan2021application}. Metabolic flux, governed by mass action kinetics, measures the rates at which metabolites are produced or consumed within a metabolic network \cite{wiechert2007fluxomics,espinel2024hybrid}. While PINNs have successfully modeled these systems, obtaining time-course data could be challenging especially when involving live objects such as model organisms or humans. 

\begin{figure*}[ht] % picture
    \centering
    \includegraphics[width=0.80\linewidth]{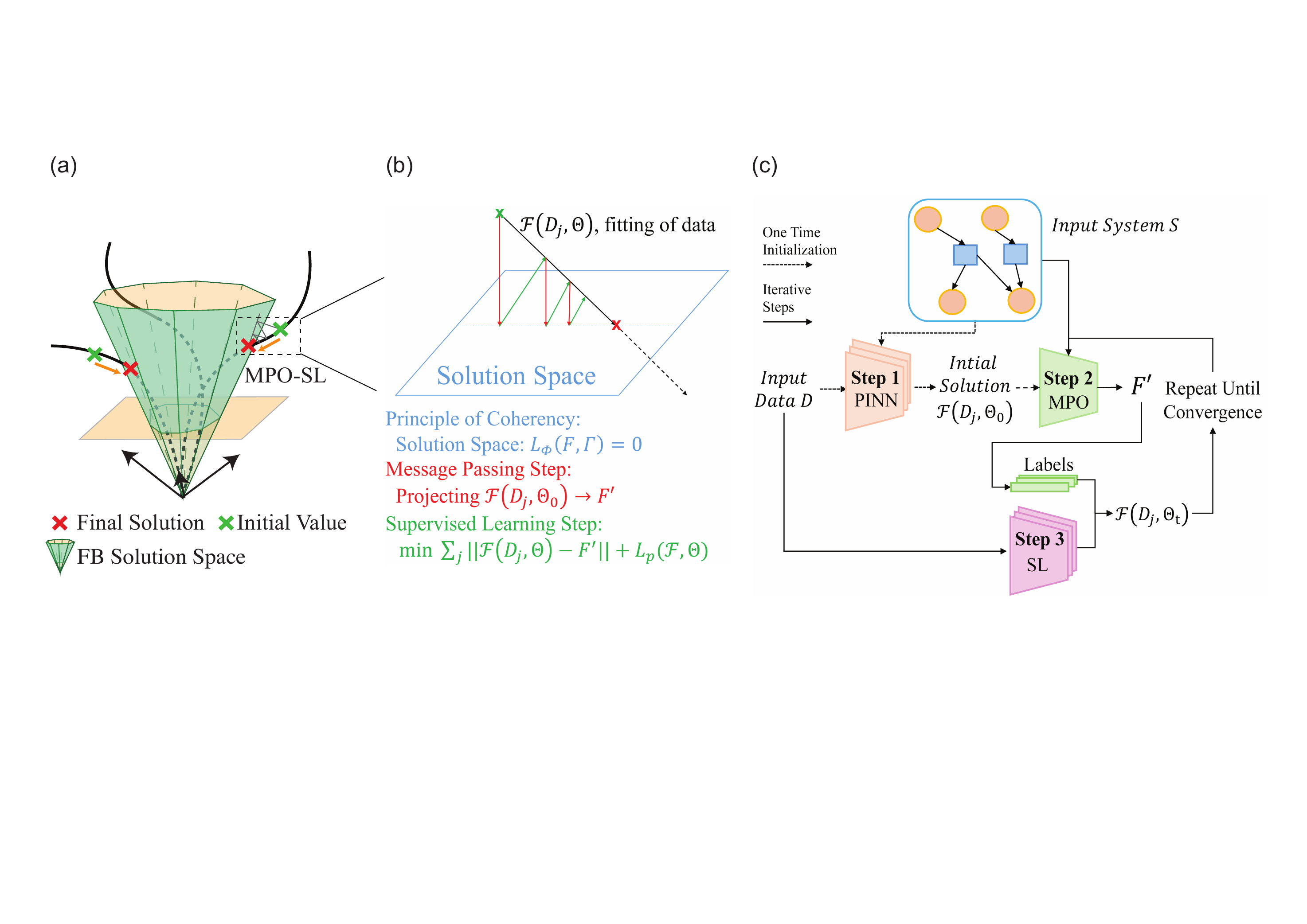}
    \caption{(a) Constrained Learning-based formulation of the flux estimation problem, (b) geometric illustration of the MPOCtrL optimization algorithm, (c) framework of the MPOCtrL algorithm.}\label{problem_statement}
\end{figure*}

We next briefly review the background of metabolic flux modeling. Flux balance analysis (FBA) has been a popular tool for modeling metabolic flux dynamics that solves the flux rate of reactions in a complex metabolic network under steady-state assumption\cite{orth2010flux,kauffman2003advances,raman2009flux}. The steady-state conditions describes that intermediate metabolites will not change over time, constraining the solution space of flux on the surface of a high dimensional polytope \cite{varma1994stoichiometric} (Figure \ref{problem_statement}a). %Over time, the method has been enhanced to incorporate genetic and regulatory information, leading to the development of more sophisticated models such as those discussed by Covert and Palsson \cite{covert2002transcriptional}, which integrate transcriptional regulatory networks with metabolic networks to simulate more complex biological behaviors. 
Recent developments have focused on integrating FBA with high-throughput genomic data\cite{lemmon2013high} to enable a data-driven flux analysis \cite{henry2010high}. For example, Wagner et al. \cite{wagner2021metabolic} introduced \textbf{Compass}, and Alghamdi, et al. \cite{alghamdi2021graph} introduced a graph neural network model called \textbf{scFEA}, both of which leverages single-cell RNA-seq (scRNA-seq) data and FBA to infer metabolic states at the cellular level. %This tool is significant because it allows researchers to pinpoint metabolic pathways that could be manipulated to alter cellular functions, offering potential therapeutic targets for autoimmune diseases. Compass effectively maps out a metabolic switch similar to that observed between different cell types, underscoring its role in cellular function and disease progression. The study showcases the potential of integrating computational modeling with biological data to understand complex metabolic networks in unprecedented detail. 
%Similarly, the work by Alghamdi, et al. \cite{alghamdi2021graph} introduced a graph neural network model called \textbf{scFEA} to estimate cell-wise metabolic activity using scRNA-seq data. %The scFEA model incorporates the principles of PINNs to enhance the estimation of reaction rates within biological networks. This approach utilizes several specially designed neural networks, leveraging established biological laws to construct a tailored loss function for learning process. This loss function essentially guides the neural networks to adhere closely to biological realities, thus ensuring that the predicted reaction rates are not only scientifically plausible but also accurate. By integrating these laws directly into the model's learning process, scFEA significantly enhances the resolution and practical utility of FBA in systems biology. This method exemplifies how modern machine learning techniques can be adapted to complex biological contexts, potentially leading to more precise and insightful analyses in systems biology and related fields. 
While Compass and scFEA estimate flux dynamics using static data, they face significant challenges due to their reliance on comprehensive and high-quality genomic datasets, which are often sparse and noisy.  In addition, both methods lack robust feature selection strategies to select relevant biological features from complex networks. Moreover, scFEA’s neural network-based approach suffers from convergence and stability issues.
%Compass, similarly, does not incorporate a feature selection mechanism, which can lead to the oversimplification of metabolic models and fail to capture the full variability observed in cell populations. 
These limitations highlight the need for a robust and interpretable method that could leverage the power of PINNs to handle non-time course data.
%As research progresses, integrating FBA with other computational methods and experimental data continues to be a promising direction, aiming to overcome existing limitations and enhance the predictive capabilities of metabolic models. These advances are likely to play a crucial role in biotechnology, medicine, and ecology, providing deeper insights into metabolic diseases, engineering microbial production systems, and understanding ecosystem dynamics.

\begin{figure*} % picture
    \centering
    \includegraphics[width=1.0\linewidth]{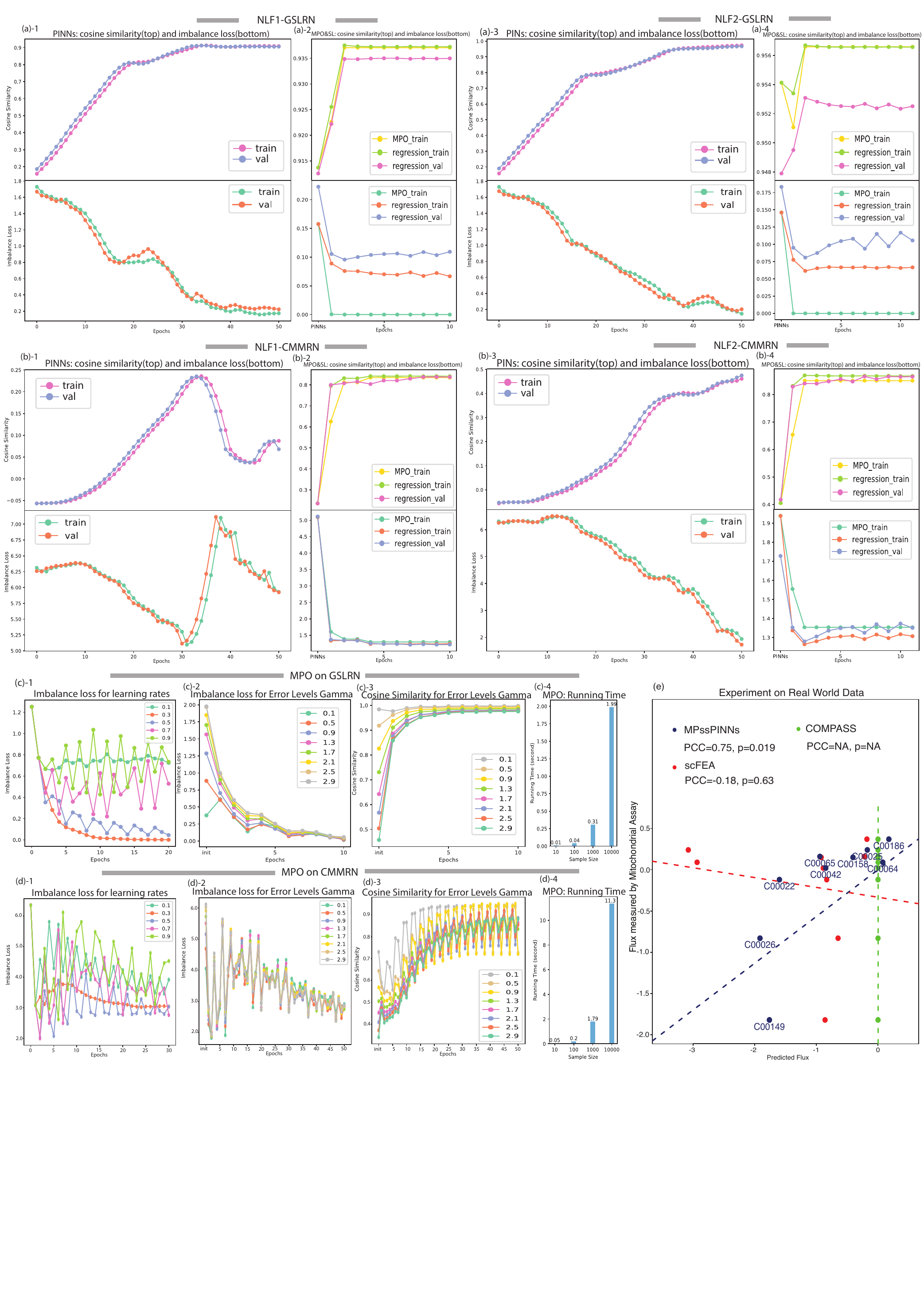}
    \caption{Experiments of MPOCtrL and MPO on Synthetic Data and Real-world Data}\label{MPOCtrL_experiments_main_figs}
\end{figure*}

\section{Mathematical consideration and method formulations}

\subsection{Notations and preliminaries}\label{notations_and_preliminaries}
\textbf{General notations}. Let $X=\{x_1,...,x_n\}$ represent a vector. Denote $X_i$ as the $i$-th element of $X$. Denote $Y^{M \times N}$ as a matrix with $M$ rows and $N$ columns, $*$ as matrix product, $\cdot$ as inner product, and $Y^T$ as the transpose of $Y$. Denote $Y_{i,j}$ as the ${(i,j)}$-th entry of $Y$, and $Y_{i,\cdot}$ and $Y_{\cdot,j}$ as the $i$-th row and $j$-th column of $Y$.  Denote $\{ Y^k \}_{k=1...K}^{M_k\times N_k}$ as a set of matrices, in which the matrix $Y_k$ has $M_k$ rows and $N_k$ columns. Denote $G=(V, E)$ as a graph with a node set $V$ and an edge set $E$. 

\textbf{Dynamic system and PINN}. This study demonstrates the concept of Constrained Learning through the lens of ODE-based dynamic systems. We represent a dynamic system using the general tuple definition: $S\overset{\Delta}{=}\{X, F, T\}$, where $X=\{x_1,...,x_n\}$ is a set of variables that take values on a finite space $\mathcal{X}=\{\mathcal{X}_1,...,\mathcal{X}_n \} \subseteq \mathbb{R}^n$, $T$ denotes time, and $F$ is the set of functions that characterizes the change of $X$ over $T$, $F \overset{\Delta}{=} \{\dv{x_i}{t}|i=1,...,n\}: U \subseteq (\mathcal{X} \times T)\rightarrow \mathcal{X} $. A conventional PINN-based solution of $S$ approximates $F$ by fitting the dynamic changes of $X$ over time using observed data $D_{time}=\{\{X_{t_1},t_1\},...,\{X_{t_m},t_m\}\}\subset \mathcal{X}\times T$, where $t_1,...,t_m$ are the time of observations that can be either continuous or discrete. In contrast, our study addresses the approximation of $F$ under two challenging conditions (1) the time information $t$ is either absent or too scarce in $D$ to support a supervised fitting of $F$, and (2) $X$ is not fully observed in $D$, as defined below.
\begin{definition}\label{definition_1}
  \textbf{PINN for dynamic systems using non-time course and partially observed data}(Problem Definition). For a dynamic system $S=\{X=\{x_1,...,x_n\}, F, T\}$ and non-time course data 
$D$ containing partial observations of $X$ from $m$ samples, $D=\{D_1,...,D_m\}\subset \{\mathcal{X}\cup\mathcal{L}\}$, the goal of this PINN problem is to identify functions $\mathcal{F}_i$ to approximate $\dv{x_i}{t}$, $\forall i=1,...,n$; and obtain the functional values of $\mathcal{F}_i$ associated with each sample in $D$, i.e., $\mathcal{F}_i(D_j)\sim\dv{x_i}{t}{(j)}$, $\forall j=1,...,m$. 
\end{definition}

By Definition \ref{definition_1}, $D$ doesn't contain effective time information. For instance, in traffic flow problems, car counts collected hourly have time gaps too large to estimate real-time traffic flow accurately; in biological flux estimation problem, biological reaction rate over time can not be collected at all. Consequently, these observations can only be viewed as independent samples. In addition, $\mathcal{L}$ is a latent space that indicates $E$ could be partially or entirely unobserved. For example, in traffic flow approximation, one might collect roadside noise or dust levels instead of actual car counts, leading to incomplete or indirect observations.
 
\textbf{Directed Factor Graph-based representation of dynamic systems}. To solve the dynamic system $S=\{X, F, T\}$ using non-time course data, we introduce a Directed Factor Graph (DFG) to represent $S$. Denote a DFG as $G^{DF}=(V^{DF}, E^{DF})$, where $V=\{V_{fa},V_{va}\}$ includes two types of nodes, namely $Factors\ (V_{fa})$ and $Variables\ (V_{va})$. In $G^{DF}$, a factor node can be only linked by a variable node, and a variable node can be only linked by a factor node. Each edge has a direction. Thus, $E^{DF}$ consists of all edges from factor nodes to variable nodes, i.e., $E_{V_{fa}\rightarrow V_{va}}$, as well as all edges from variable nodes to factor nodes, i.e., $E_{V_{va}\rightarrow V_{fa}}$. To define $S$ over $G^{DF}$, we first let $V_{fa}\overset{\Delta}{=}X=\{x_1,...,x_n\}$. We further define a set of functions $f_k: U_k \subseteq \mathcal{X} \rightarrow U_k' \subseteq\mathcal{X}, U_k \cap U_k' = \emptyset, k=1,...,K$, i.e. each $f_k$ takes two non-overlapping subsets of $V_{fa}$ as its input and output, which are denoted as $V^{in}_{f_k}$ and $V^{out}_{f_k}$. Then we define $V_{va}\overset{\Delta}{=}\{f_k\}$. For a given $S=\{X, F, T\}$, we can further derive that there always exists a set of $f_k$ as defined above, for any $\dv{x_i}{t}\in F: U \subseteq (\mathcal{X})\rightarrow \mathcal{X}_k$, $\dv{x_i}{t}=\sum_{k|x_i\in V^{out}_{f_k} } \gamma_{ik} f_k + \sum_{k'|x_i\in V^{in}_{f_{k'}} } \gamma_{ik'}f_{k'}$, where $\gamma_{ik}$ and $\gamma_{ik'}$ are bounded positive and negative weights that could be pre-computed based on the physical property of $S$. Because the input and output sets of $f_k$ are non-overlapping, $\gamma_{ik}$ and $\gamma_{ik'}$ can be stored by one $n\times K$ real matrix $\Gamma$, in which $\Gamma_{ik}<0$ if $x_i\in V^{in}_{f_k}$,  $\Gamma_{ik}>0$ if $x_i\in V^{out}_{f_k}$, and $\Gamma_{ik}=0$ if $x_i$ if not adjacent to $f_k$. Then the edge set $E^{DF}=\{E_{V_{fa}\rightarrow V_{va}},E_{V_{va}\rightarrow V_{fa}}\}$ can naturally be defined by $\{V^{in}_{f_k}\rightarrow f_k,f_k\rightarrow V^{out}_{f_k}\}$. Thus, $S$ could be represented by $G^{DF}$ as $X=V_{fa}$ and $F=\Gamma* V_{va}$. We call $f_k$ as \textbf{message} functions. The impact of $f_k$ on $\dv{x_i}{t}$ is controlled by the physical weight $\gamma_{ik}$. 

\textbf{ODEs in $S$, message over $G^{DF}$, and physical constraints-informed learning}. For the PINN problem described in \textbf{Definition 1}, observations in $D\subset \{\mathcal{X}\cup\mathcal{L}\}$ can be viewed as attributes on $G^{DF}$ by assigning $D\cap V_{fa_i} \in \mathcal{X}_i$ to the $i^{th}$ factor node $V_{fa_i}$ and related features in $\mathcal{L}$ to the $k^{th}$ variable node $V_{va_k}$, denoted as $D\cap V_{va_k}\in \mathcal{L}$. Because $\{\dv{x_1(j)}{t},...,\dv{x_n(j)}{t}\}^T= \Gamma* \{f_1(D_j),...,f_k(D_j)\}^T$, identifying $\mathcal{F}_i(D_j)\sim\dv{x_i(j)}{t}$ is equivalent to identifying $\{f_k\}$, here the input of each $f_k(D_j\cap V_{va_k},D_j\cap V_{fa_i}|i\in \{V^{in}_{f_k}, V^{out}_{f_k}\}$ is the subset of $D_j$ attributed to $V_{va_k}$ and $\{V^{in}_{f_k}, V^{out}_{f_k}\})$. Noted, $F =\dv{X}{t}: U \subseteq (\mathcal{X} \times T)\rightarrow \mathcal{X} $. When time course information in $D$ is available and $E$ is fully observed by $D$, denoted as $D_{x_i}^t\overset{\Delta}{=} D^{t}\cap V_{fa_i}$, a supervised framework can be utilized to identify $f_k$ by fitting $\{D_{x_1}^{t+1}-D_{x_1}^{t},...,D_{x_n}^{t+1}-D_{x_n}^{t}\}^T \sim \Gamma*\{f_1(D_j^t),...,f_k(D_j^t)\}^T \cdot \Delta(t+1,t)$. Thus, the \textbf{message} functions $\{f_k\}$ characterize the \textbf{physical flow} from $V^{in}_{f_k}$ to $V^{out}_{f_k}$ via $V_{va_k}$ per unit time.

For the problem described in \textbf{Definition \ref{definition_1}}, when time information is unavailable, or $E$ is not fully observed in $D$, additional constraints are needed to approximate $\{f_k\}$. Real-world dynamic systems follow physical laws and context-specific system properties, such as the law of mass conservation and thermodynamics. These physical laws and system-specific properties constrain the solution space of $\{f_k\}$, and can be encoded and generalized as a loss function \(L_{\Phi}(\{f_k\},\Gamma)\), which has a smaller value when $\{f_k\}$ is more coherent to the physical laws and system-specific properties. And specifically, \(L_{\Phi}(\{f_k\},\Gamma)=0\) when $\{f_k\}$ satisfy all the physical laws and system-specific rules. The derivation of $L_{\Phi}$ for a few different types of dynamic systems and PINN problems was illustrated in APPENDIX.

\subsection{Constrained Learning to solve PINN with non-time course and partially observed data} 
By introducing \(L_{\Phi}\), we develop a new learning framework namely \textbf{Constrained Learning} that approximates $\{f_k(D)\}$ by leveraging the coherency to the physical laws and system-specific properties and the goodness of fitting to the data. Enlightened by Ma et al.\cite{ma2022principles}, we summarized two necessary computational principles, namely: (i) \textbf{\textit{principle of coherency}}: an intelligent system-based approximation should seek to maximize the coherency to the physical laws of the system on the observed data, (ii) \textbf{\textit{principle of parsimony}}: an intelligent system needs to be simple and structured when approximating a dynamic system. The principle of coherency describes that an approximation of a dynamic system should maximally represent (1) the physical laws, (2) context-specific dynamic properties such as functional forms of the ODEs, and (3) other prior knowledge. In APPENDIX, we summarized ODE-based dynamic systems into sub-categories and provided each category's mathematical forms of $L_{\Phi}$. Figure \ref{problem_statement}a showcases Constrained Learning on the flux estimation problem, in which the pyramid is the solution space constrained by the physical laws of the flux balance system and the strings are functions that could be fitted by data. The principle of parsimony describes that a good approximation of $\{f_k(D)\}$ should have a relatively simple functional form, to avoid overfitting.

\begin{definition}\label{definition_2}  
\textbf{Constrained Learning} For a dynamic system $S=\{X=\{x_1,...,x_n\}, F=\dv{X}{t}, T\}$ and observed data $D$, denote the DFG representation of $S$ by $G^{DF}=\{\{V_{fa}=X,V_{va}=\{f_k\}\}, E^{DF}\}$ and $\Gamma^{n\times K}$ as the system specific weight matrix s.t. $F=\Gamma* \{f_1,...,f_k\}^T$, the goal of Constrained Learning is to the PINN problem described in \textbf{Definition \ref{definition_1}} by identifying (i) physics-informed loss term $L_{\Phi}$ that characterizes the physical laws and system properties of $S$ s.t. \(L_{\Phi}(\{f_k\},\Gamma)=0\), and (ii) functions $\mathcal{F}=\{\mathcal{F}_k\}$ and parameters \(\Theta=\{\Theta_k\}\) that takes $D$ as the input, such that $\{\mathcal{F}_k(D,\Theta_k)\}$ forms a good approximation to \(\{f_k(D)\}\).Specifically, with having $L_{\Phi}$ defined, $\mathcal{F}$ and $\Theta$ could be identified by minimizing the following loss function
\[
L_{\Phi} (\mathcal{F}(D,\Theta),\Gamma)+ L_p(\mathcal{F}, \Theta) \quad \textcircled{1}
\]
, here the first term $L_{\Phi}$ regularize if $\mathcal{F}$ is coherent with the physical laws and context properties of $S$. The second term \(L_p(\mathcal{F}, \Theta)\) regularize the parsimony property of \(\mathcal{F}\) and \(\Theta\). 
Then $F$ could be approximated by $F=\Gamma* \{\mathcal{F}_1(D,\Theta_1),...,\mathcal{F}_k(D,\Theta_K)\}^T$.
\end{definition}

Noted, the derivation of $L_\Phi$ is system and problem-specific. More prior knowledge or assumptions of the system provides a higher strength of $L_\Phi$ in regularizing the solution space of $\mathcal{F}$. In section \textbf{\ref{new_optimization_strategy}}, we first provide a general optimization framework to solve the general Constrained Learning problem. In sections \textbf{\ref{estimating_flux}} and \textbf{\ref{experiments}}, we further demonstrate the utility of the Constrained Learning framework by applying it to solve the data-driven flux analysis problem.

\subsection{A new optimization strategy to solve the general Constrained Learning problem}\label{new_optimization_strategy}
Minimization of the loss function defined in \textcircled{1} is non-trivial because it needs to search over the functional space of \(\mathcal{F}\) and parameter space of \(\Theta\). Noted, minimizing \textcircled{1} neither falls into the paradigm of supervised learning nor unsupervised learning. The parsimony loss $L_p(\mathcal{F}, \Theta)$ cannot be directly co-optimized with coherency loss $L_{\Phi} (\mathcal{F}(D,\Theta),\Gamma)$ by existing optimization approaches such as gradient descent \cite{ahn2022understanding,cheridito2022proof}. Noted, $\mathcal{F}$ approximates $f_k$, which are variable nodes over a DFG and each $f_k$ can be viewed as a function that transfers messages or physical flow among factor nodes, then $L_{\Phi} (\mathcal{F}(D,\Theta),\Gamma)$ defines a certain dependency of $\mathcal{F}(D,\Theta),\Gamma)$ over the DFG, and its minimal value can be viewed as the state of the most balanced message flow among the factor nodes, for which state of the most balanced message flow over the DFG can be achieved by message passing\cite{ihler2005loopy,kuck2020belief}. Continually using the formulations given in \textbf{Definition \ref{definition_1} and \ref{definition_2}}, we developed this Message Passing Optimization-based Constrained Learning \textbf{MPOCtrL}, a novel self-supervised method to minimize \textcircled{1}, as illustrated in Figure \ref{problem_statement} b,c and detailed below:

Building upon the previously introduced loss functions, MPOCtrL leverages neural networks and incorporates critical domain knowledge through a combined loss framework.  The core principle is to guide the learning process by explicitly enforcing physical laws or constraints, enabling the network to discover underlying patterns and relationships within the data without explicit supervision.  MPOCtrL achieves this through the synergistic combination of parsimony regularization incorporating a gating mechanism, a coherency loss reflecting physical constraints, and a novel message passing mechanism to accelerate convergence to physically plausible solutions.

MPOCtrL employs a neural network as the central learning engine, receiving input data and learning to extract relevant features and representations.  The learning process is driven by minimizing a combined loss function, as detailed in the preceding sections, comprising three key components:\\
(1) Parsimony Loss with Gating Mechanism and L2 Regularization: This component promotes simpler and more interpretable solutions.  Beyond standard L2 regularization on the network's weights to prevent overfitting, MPOCtrL incorporates a gating mechanism.  This mechanism selectively activates or deactivates neurons or even entire layers within the network, effectively controlling the complexity of the learned representations. The gating mechanism allows the network to learn which parts of its architecture are most relevant for capturing the essential features of the data, promoting sparsity and interpretability.  The parsimony loss, in conjunction with the gating mechanism and L2 regularization, encourages the network to learn generalizable features while avoiding overfitting to noise or irrelevant details.  This combined approach provides a more robust and effective way to control model complexity compared to L2 regularization alone.\\
(2) Coherency Loss: This component is the cornerstone of MPOCtrL, enforcing the physical laws governing the system. It measures the degree to which the network's output satisfies these constraints, providing essential domain knowledge to guide the network towards physically meaningful solutions.\\
(3) Message Passing Optimization: This novel component accelerates convergence by explicitly propagating information about the constraints throughout the network.  It facilitates "constraint-aware" communication, enabling different parts of the network to coordinate their learning based on global constraints.

The strength of MPOCtrL lies in the interplay of these three components. The parsimony loss promotes simplicity, the coherency loss enforces physical consistency, and the message passing loss accelerates convergence.  By minimizing this combined loss function, MPOCtrL learns to extract meaningful representations from unlabeled data while adhering to underlying physical principles. This framework offers a powerful approach for tackling challenging self-supervised problems where domain expertise can be expressed through constraints.  In the following sections, we demonstrate the effectiveness of MPOCtrL through a series of experiments, showcasing its ability to learn physically consistent representations from unlabeled data.

The message passing step is highly efficient. By caching pre-calculated messages, we optimized its time complexity to $O(nK)$ and space complexity to $O(nK)$, where $n$ and $K$ are the numbers of factor and variable nodes of the input DFG. Furthermore, MPO's global perspective ensures balanced updates across the entire network, leading to convergence more efficiently. A detailed discussion of the MPO step is given in APPENDIX section \ref{mpo}.

\subsection{Estimating flux over a network by Constrained Learning}\label{estimating_flux}
\textbf{DFG representation and physical constraints of the flux estimation problem.} As discussed in sections \textbf{\ref{introduction}} and \textbf{\ref{Data_Driven_Flux_Analysis}}, flux analysis has been broadly utilized in physics, social, and biological science, such as cash flow tracking, traffic flow prediction, electric flux, fluid dynamics, and metabolic flux analysis. To demonstrate the utility of Constrained Learning in solving the PINN problems described in \textbf{Definition \ref{definition_1}}, we apply the framework to solve the mass-carrying flux over a network using non-time course and partially observed data. We first formulate the mass-carrying flux analysis problem using the DFG-based representation of a dynamic system, as described in \textbf{\ref{notations_and_preliminaries}}. For a system $S=\{X=\{x_1,...,x_n\}, F, T\}$ that is formed by mass-carrying flux, $\{f_k\}$ is defined as the set of all mass-carrying flux between two non-overlapped subsets of variables in the system. For each $f_k$, $V^{in}_{f_{k}}\subset X$ and $V^{out}_{f_{k}}\subset X$ are the input and output set of $f_k$, and the value $f_k$ is the level of the mass transfer from $V^{in}_{f_{k}}$ to $V^{out}_{f_{k}}$. In the traffic flow prediction task, each element in $X$ is a city or an intersection, and $\{f_k\}$ is the flux of each road. In the metabolic flux estimation task, each element in $X$ is a metabolite, and $\{f_k\}$ is the flux of each chemical reaction that takes $V^{in}_{f_{k}}$ as substrates and $V^{out}_{f_{k}}$ as products. Then the DFG that represents $S$ could be derived as \(G^{DF}=\{V_{fa}={X}, V_{fa}={f_k}, E=\{V^{in}_{f_{k}} \to f_k, f_k \to V^{out}_{f_{k}}\}\} \). Here, ${f_k}$ is the to-be-approximated flux rate and $\dv{x_i}{t}=\sum_{k|x_i\in V^{out}_{f_k} } \gamma_{ik} f_k + \sum_{k'|x_i\in V^{in}_{f_{k'}} } \gamma_{ik'}f_{k'}$. In the traffic flow task, each $f_k$ has one input node and one output node, denoted as $i_k^{in}$ and $i_k^{out}$, then $\gamma_{i_k^{in}k}=-1$, $\gamma_{i_k^{out}k}=1$, and $\gamma_{ik}=0$ for other $i$. In the metabolic flux task, $\gamma_{ik}$ is the coefficient of $x_i$ in the chemical reaction $f_k$, and $\Gamma^{n\times K}$ is the stoichiometry matrix\cite{alghamdi2021graph,llaneras2008stoichiometric}. Under steady state or quasi-steady state, by the law of conservation of mass, $\dv{x_i}{t}=0$ or a small value for all intermediate variables that suggests that $\Gamma*\{f_1,...,f_K\}^T=0$ or be small. Thus, $L_\Phi$ could be defined by $L_\Phi=||\Gamma*\{f_1,...,f_K\}^T||_2$.

\textbf{Flux estimation using MPOCtrL.} To demonstrate the power of Constrained Learning in solving the flux estimation problem as described in \textbf{Definition \ref{definition_1}}, we assume the observed data is non-time course and $V_{fa}=X$ is totally unobserved, i.e., $D\cap {V_{fa}}=\emptyset$. For a given data $D$ of $m$ samples, denote $D_{kj}$ as the observation of the features assigned to the $k^{th}$ variable node $V_{va_k}=f_k$ in the $j^{th}$ sample. Following \textbf{Definition \ref{definition_2}}, Constrained Learning identifies functions \(\{\mathcal{F},\Theta_k\}=\{\mathcal{F}_k(D_{kj}, \Theta_k)\} \) to approximate $\{f_k\}$ by minimizing the following loss function:
\[
\mathcal{L} = L_\Phi(\mathcal{F}(D,\Phi),\Gamma)+ L_p(\mathcal{F}, \Theta) = \sum_{j \in {1,...,n}}||\Gamma*\{f_1(D_{1j}),...,f_K(D_{Kj})\}^T||_2 + L_p(\mathcal{F}, \Theta) 
\]
\[
= \sum_{j=1}^{m} \sum_{i=1}^{n} \left( \sum_{k|x_i\in V^{in}_{f_k}} \mathcal{F}_{k}(D_{kj},\Theta_k) - \sum_{k|x_i\in V^{out}_{f_k}}  \mathcal{F}_{k'}(D_{k'j},\Theta_k') \right)^2 + L_p(\mathcal{F}, \Theta)  \quad \textcircled{2}
\]
The above derivations enable the implementation of Constrained Learning on the flux estimation problem. We name this new PINN method as \textbf{Message Passing Optimization-based Constrain Learning} (\textbf{MPOCtrL}) (Algorithm \ref{algorithm-MPOCtrL} in APPENDIX). MPOCtrL takes a DFG $G^{DF}$ and observed data $D$ as input and utilizes the biological laws to compute $\mathcal{F}(D,\Theta)$ that minimizes $\mathcal{L}$ (see details in APPENDIX). We also developed the MPO Algorithm for this specific task (Algorithm \ref{algorithm-mpo}). To evaluate the performance of MPOCtrL, we compare it with baseline methods on synthetic and real-world data in \textbf{Section 4}.

\section{Experiments}\label{experiments}
\subsection{Experimental setup}
In this section, we comprehensively evaluated the performance of MPOCtrL and its sub-algorithms on both synthetic and real-world datasets. We assessed: (i) the overall performance of MPOCtrL compared with state-of-the-art (SOTA) methods including scFEA and Compass, and (ii) the effectiveness, robustness and efficiency of the MPO algorithm specifically. 

\textbf{Overall design.} For each experiment, the input to each method includes a network, which is a directed factor graph, and (non-time course) observations of attributes of the network, i.e., attributes of the variable nodes. An output of flux over the network is expected from each method. For synthetic experiments, we utilized three real-world biological reaction networks and one highly complex synthetic network. Importantly, none of them contains self-loops. For each network, two sets of observation data were simulated. This resulted in a total of eight synthetic input scenarios, each simulated multiple times. For real-world experiments, we used one real-world dataset, which contains one context-specific biological network, and real-world observations of network attributes, i.e., all the variable nodes. Our method was mainly implemented by Python, PyTorch\cite{paszke2019pytorch}. The random seed for all potential random steps is set to 42 in all experiments. And the experiments were implemented on a computer server equipped with 256 GB of memory and two 64-core, 2.25 GHz, 225-watt AMD EPYC 7742 processors.

\textbf{Evaluation Metrics.} On synthetic experiments, we used two evaluation metrics: (1) the mean of sample-wise cosine similarity between predicted and true network flux; and (2) imbalance loss which measures the overall differences between the in- and out-flux for all metabolites. Note that under the stringent flux balance condition, the in-flux equals to out-flux. For real-world datasets, (3) we used experimentally measured flux as ground truth to assess the predicted flux by each method.

\textbf{Evaluated Networks} We tested our method on four real-world biological reaction networks, each containing metabolites (factors) and reactions (variables) that are crucial to cancer treatment\cite{zhang2023fluxestimator}. Additionally, we evaluated our approach on a highly complex synthetic directed factor graph. The details of each network are as follows: (1) Antigen Presentation Reaction Network (APRN) includes 10 metabolites, and 11 reactions, as shown in Fig \ref{aprn} in APPENDIX. (2)Glutamine Subcellular Localization Reaction Network (GSLRN), contains 6 metabolites and 10 reactions, as shown in Fig \ref{gslrn}. (3) Central Metabolic Map Reaction Network (CMMRN) was encapsulated in scFEA\cite{alghamdi2021graph} comprises 66 metabolites, 159 reactions and 3 cycles, as shown in fig \ref{cmmrn}. (4) Glutamine Glucose Subcellular Localization Reaction Network (GGSLRN), contains 23 metabolites and 42 reactions, as shown in fig \ref{ggslrn}, This network was exclusively applied to real-world data due to its relevance and applicability to the specific disease being studied. (5) Synthetic Directed Factor Graph (SDFG), a more complicated synthetic network. This synthetic network has 204 factors, 453 variables, and 18 cycles, as shown in \ref{sdfg}. Detailed information of these networks is given in APPENDIX. 

For the real-world biological reaction networks, the count and name of genes (features) in the reactions (variables) are also known. Then we converted metabolites into factors and reactions into variables to match the $G^{DF}$ format for the experiments. Each factor represents a linear relationship between its parent and child variables. Each variable contains observation data in matrix format, where rows represent samples and columns represent features, along with the estimations for the samples we need to estimate. Here, in the biological cases, the samples could be cells, features could be genes involved in the reaction, the estimations could be the fluxes for the cells in the reaction.

\begin{table}
    \centering
    \caption{Experiment on Synthetic Observation Data.}\label{table_1_methods}
     \includegraphics[width=0.65\linewidth]{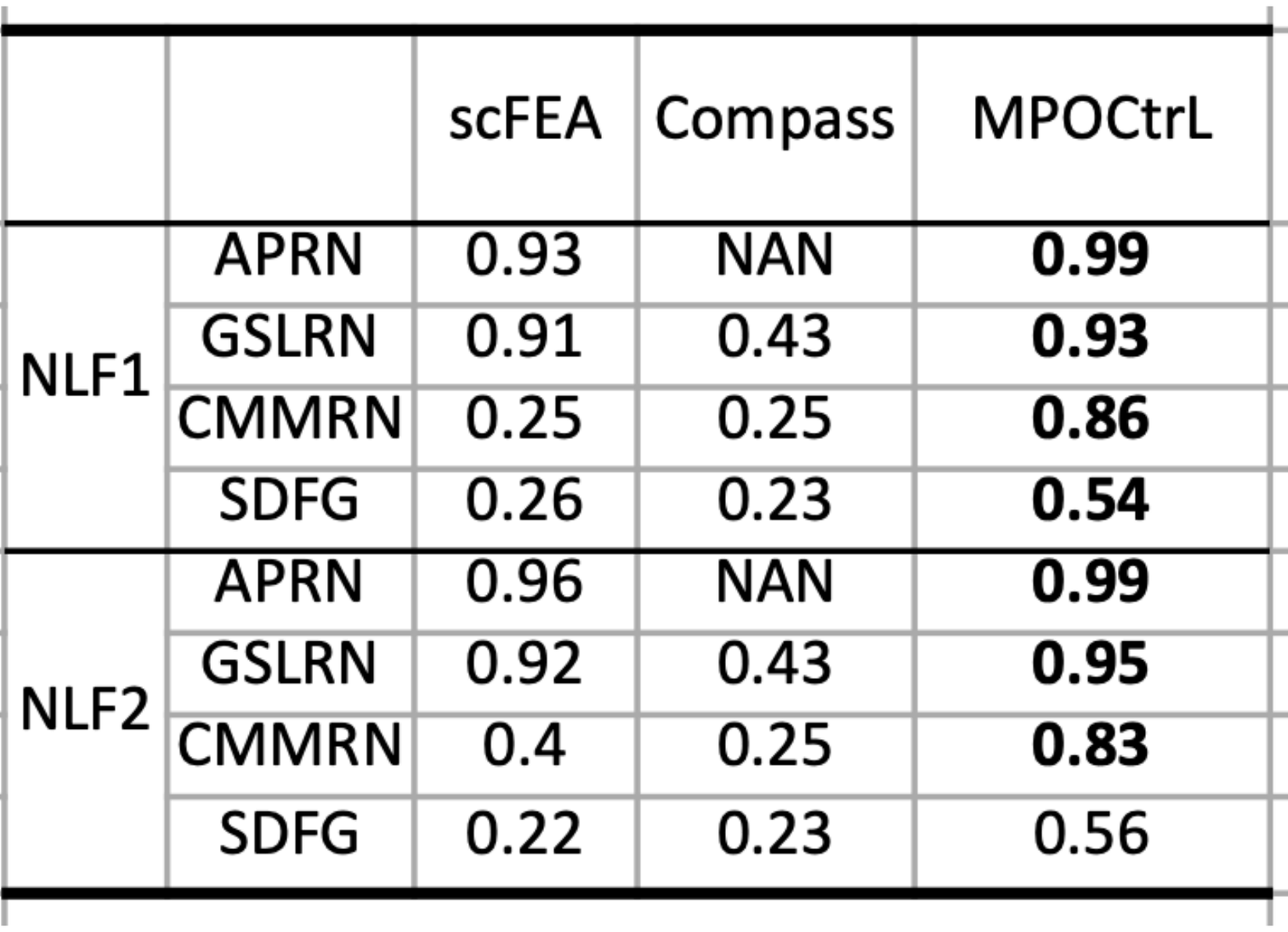}
\end{table}

\textbf{Generation of Synthetic Observation Data.} 
%For a given network, we link the flux of each reaction to the values of associated network attributes with two different non-linear formulas (NLFs): (1) NLF1, $Y_{j,l} = a \times (\sum_{i\in Z_l} D_{j,z})^2 + b \times (\sum_{z\in Z_l} D_{j,z})$; and (2) NLF2,  $Y_{j,l} = \exp\{a \times (\sum_{i\in Z_l} D_{j,z})^2 + b \times (\sum_{i\in Z_l} D_{j,z})\}$. Here $Y_{j,l}$ denotes the true flux for the $l$-th reaction in the $j$-th sample; $D_{j,z}$ denotes the $z$-th attribute value in the $j$-th sample; $Z_l$ indicates all those node attributes in $D_{j,.}$ that is associated with the $l$-th reaction; $a,b$ are randomly generated constants in interval (0,10). We start by simulating the reaction flux for multiple samples, i.e., $Y$, and solve attribute values $D$ in a backward fashion according to the NLFs. See more details in APPENDIX A.3.2.

To simulate synthetic observation datasets, we start by simulating the network flux; the observation of network attributes were simulated in a backwards fashion by solving a non-linear equation linking the network flux and the network attributes. Two such non-linear formulas (N-LFs) were used: (1) NLF1, $W_k = a \times (\sum_{i=1}^{n^k} X_{k,i})^2 + b \times (\sum_{i=1}^{n^k} X_{k,i})$ and (2) NLF2, $W_k = \exp\{a \times (\sum_{i=1}^{n^k} X_{k,i})^2 + b \times (\sum_{i=1}^{n^k} X_{k,i})\}$, where $W_k$ is the weight for the $k_{th}$ variable, $X_{k,:}$ and $n^k$ mean the $k_{th}$ unknown observation data and number of features corresponding to the $k_{th}$ variable, $X_{k,i}$ is the $i_{th}$ feature in $k_{th}$ observation. Here, $a,b$ are randomly generated Coefficients range from (0,10) and $W$ denotes the simulated known balanced weights for variables. To generate the target $W$, we randomly generate a set of positive numbers and input them into the MPO algorithm to obtain balanced initial values on a directed factor graph. This process ensures that the initial values are adjusted appropriately within the graph structure to meet the balancing criteria. To simulate the synthetic observation data $X_{k,:}$, we firstly randomly generate $n^k$ positive numbers and set $20\%$ of them to zero to simulate data sparsity, then normalized by dividing them by sum of all numbers to ensure their sum $\sum_{i=1}^{n^k} X_{k,i}$ equals one. Using the quadratic formula, we solved for unknown $\sum_{i=1}^{n^k} X_{k,i}$ and multiplied it by the previous random positive numbers, ensuring the correct non-linear relationship between $W_k$ and $X_{k,:}$. This approach allowed us to easily and meticulously control the synthetic data's complexity and variability, making it a robust tool for evaluating our method's performance under controlled conditions.

\textbf{Real-World Data.} One real-world scRNA-seq data was retrieved from the NCBI GEO database with accession ID GSE173433. The dataset measures single-cell transcriptomic profiles of the Pa03c pancreatic cancer cell line under APEX1 inhibition and control. Matched mitochondrial assay data that measures the reaction flux of nine metabolites was provided.

\subsection{Experiments on Synthetic Observation Data}
For each variable in each tested network (represented as a DFG), we simulated 500 samples using each of the two NLFs. All variables in the DFG have the same number of samples with identical sample names, but the number of observable attributes per variable may vary, and the groups of attributes for different variables may overlap. We have eight synthetic observation datasets in total: NLF1-APRN, NLF2-APRN, NLF1-GSLRN, NLF2-GSLRN, NLF1-CMMRN, NLF2-CMMRN, NLF1-FAKE, NLF2-FAKE. We divided the dataset into three subsets: $60\%$ for training, $20\%$ for validation, and $20\%$ for testing.

The MPOCtrL architecture employs an ensemble of neural network groups, each designed to process input data with varying dimensionalities. Each group comprises a series of fully connected neural networks, where the architecture of each individual network within a group is consistent but the input size can differ across groups. Specifically, each network consists of three hidden layers with sizes $2\times input size$, $4\times input size$, and $8\times input size$, respectively, followed by a gating mechanism, leaky ReLU activation, and dropout (rate 0.5).  Training utilizes the Adam optimizer with L2 regularization, learning rate 0.05, and the MPO component is invoked every 10 epochs to accelerate convergence to physically consistent solutions. The outputs from these neural network groups are then collected to calculate the flux balance loss. Figure \ref{MPOCtrL_experiments_main_figs} demonstrates the robustness and accuracy of MPOCtrL and sub-method MPO on both the training and validation datasets. In Figure \ref{MPOCtrL_experiments_main_figs} (a-1,a-3 and b-1,b-3), their top and bottom figure panels show the mean of sample-wise cosine similarities between ground truth and predictions, and imbalance loss over epochs for synthetic datasets NLF1-GSLRN, NLF2-GSLRN, NLF1-CMMRN, and NLF2-CMMRN. To avoid zero solutions and over-fitting, we applied an early stopping strategy.

%, respectively for the first step ssPINNs. When the network is simple, such as GSLRN, the first step, ssPINNs could steadily generate higher cosine similarities of around 0.9 and lower imbalance losses of about 0.2 over epochs. However, for more complex networks like CMMRN, which contains more factors, variables, and loops, ssPINNs struggle. This issue arises because ssPINNs rely on a single total loss combined from all neural networks, which is insufficient to effectively guide the learning process in complex networks. Despite this, figures (a-2,a-4 and b-2,b-4) top and bottom demonstrate that the second step, MPO, and the third step, supervised learning, can still achieve convergence based on the initial estimations provided by ssPINNs. To avoid zero solutions and over-fitting, we applied an early stopping strategy. For more detailed experiments involving additional learning epochs on all eight synthetic observations, please refer to Figure \ref{MPOCtrL_appendix} in APPENDIX. These figures illustrate how well our approach performs across different complexities of data, ensuring the model's accuracy and robustness. 

Figures \ref{MPOCtrL_experiments_main_figs} (c1-c4) and (d1-d4) show MPO's sensitivity to learning rate, robustness to added noise, accuracy and running time. Additionally, Figure \ref{mpo_vs_brw_appendix} in APPENDIX compares MPO with an existing weight balancing method, BRW, across various metrics: cosine similarities between ground truth and predictions, imbalance loss, and running time. To test MPO's robustness against added noise, we input a noised version of $D$ to MPO, denoted as $\hat D$, where $\hat D =\frac{D}{norm(D)}+Error$, and $Error= \gamma \times \sum_{i=1}^{3} a_i \times v_i + \epsilon$. The added noises ensure that the inputs had controlled imbalance loss and cosine similarity that the targets deviated from their original geometrical space. Here, $v_i$ is an orthogonal vector to $D$, $\gamma = \{0.1,0.5,0.9,1.3,1.7,2.1,2.5,2.9\}$ represents different error weight levels to the orthogonal vectors, $a_i$ is a random number ranging from (0,1), $\epsilon$ is a constant number 0.1. This deviation was measured by the cosine similarity between the targets and the methods' outputs. MPO outperformed BRW with higher cosine similarities in all the datasets, ranging from simple to complex networks. BRW simply calculates the sum of weights for parent variables, then assigns these weights to the child variables by dividing the sum by the number of child variables and adding the original weights of the child variables. This approach does not account for the varying importance of variables and factors within the input network (DFG), whereas MPO considers these differences by involving neighbors' imbalance levels. We did not compare MPO with linear system-solving methods like CPLEX because these methods are not data-driven. When the linear systems are fixed, they produce identical solutions for different samples, failing to capture the sample-wise variations in the data. For more detailed experiments on other datasets, please refer to Figure \ref{mpo_vs_brw_appendix} in APPENDIX. These figures demonstrate the robustness, accuracy, and efficiency of MPO.

Table \ref{table_1_methods} presents a performance comparison between scFEA, Compass, and our method on the testing datasets. We measure their performance using the mean of sample-wise cosine similarity between ground truth and predictions. All methods achieved over 0.93 cosine similarity on synthetic NLF1-APRN and NLF2-APRN datasets, except for Compass, which produced all-zero results. The APRN network is relatively simple, and due to its architecture, all nodes' weights should be the same. For NLF1-GSLRN and NLF2-GSLRN, scFEA and our method achieved cosine similarity over 0.91, respectively. In contrast, Compass only achieved a cosine similarity of 0.43. Compass relies heavily on non-data-driven balancing methods and tends to output very similar weights across samples when the input network is fixed. CMMRN and SDFG are complex networks, with three and eighteen cycles respectively, making the problem more challenging. In scFEA, the learning strategy, which relies on a single total loss combining multiple neural networks, struggles to guide the learning process effectively when the input network is complex and contain cycles. As a result, scFEA produces low cosine similarity results. Compass faces a similar issue, often assigning very similar weights across samples, resulting in cosine similarities around 0.25. However, our method achieved over 0.83 cosine similarity on NLF1-CMMRN and NLF2-CMMRN, and over 0.54 on NLF1-SDFG and NLF2-SDFG. These results are two to three times better than the current state-of-the-art methods, demonstrating the effectiveness of our approach even in complex networks. This table highlights how our approach performs relative to established methods across various regression models and networks, demonstrating the correctness of our methodology.

\subsection{Experiments on Real-world Observation Data}

GSE173433 offers scRNA-seq and mitochondrial assay data that measures the reaction activity of nine metabolites of Pa03c cells under APEX1 inhibition and control conditions. We analyzed the scRNA-seq data by using MPOCtrL-LightGBM, scFEA, and COMPASS to predict the metabolic flux in APEX1 inhibition and control cells against the GGSLRN network. We computed the \textbf{changes in the predicted metabolic flux} of the nine metabolites between the APEX1 inhibition and control cells by the three methods and compared them with \textbf{experimentally observed flux changes} of the nine metabolites between APEX1 inhibition and control conditions. We evaluated the consistency between the predicted and experimentally flux changes using Pearson Correlation Coefficients (PCC) and its $p$-value tested by Student's t-test. A PCC=0.75 ($p$=0.019) was observed for the MPOCtrL predictions while the prediction of scFEA has a negative correlation with experimental observations and COMPASS predicted no flux changes for all metabolites despite the non-zero flux changes have been observed experimentally, as shown in Figure \ref{MPOCtrL_experiments_main_figs} (e).

\section{Conclusion}

\label{conclusion}
In this study, we developed a new learning framework named Constrained Learning and a new optimization method to solve PINN when the input data is non-time course and partially observed. We further developed a Constrained Learning method, named MPOCtrL, to solve the flux estimation problem. Benchmarking on synthetic and real-world data demonstrated the feasibility of Constrained Learning and the accuracy and robustness of MPOCtrL.
% bib
\bibliographystyle{abbrv}
\bibliography{main}

%%%%%%%%%%%%%%%%%%%%%%%%%%%%%%%%%%%%%%%%%%%%%%%%%%%%%%%%%%%%

\appendix
\setcounter{figure}{0}
\counterwithin*{figure}{part}
\renewcommand{\thefigure}{A.\arabic{figure}}

\section{Appendix / supplemental material}

\subsection{Constained Learning based solution of Flux Estimation Problem}
\label{method_MPOCtrL}
\subsubsection{MPOCtrL Framework}
In this study, we present a novel methodological framework designed to refine computational analysis, leveraging a sequence of advanced computational techniques. 

Our approach initiates by generating initial solutions through a set of self-supervised physics-informed neural networks (ssPINNs). These NNs are guided mainly by our self-supervised loss function inspired by PINNs\cite{bischof2021multi,cai2021physics,wang2022and} or scFEA \cite{alghamdi2021graph}. The loss function consists of one main component: the imbalance loss which also referred to as biological law loss, ensures that the neural network's predictions conform to established biological laws, reinforcing the scientific validity of the model. The loss function effectively guides the neural networks to balance adherence to scientific principles with sensitivity to real-world data patterns, resulting in a robust and predictive modeling approach. We integrate the DFG structure with observational data to facilitate the self-supervised learning process. This method effectively combines theoretical frameworks and practical data to enhance learning accuracy and efficiency. The advantages of this step lie in its integration of biological knowledge and data-driven learning, enabling accurate metabolic flux estimation. This initial step produces the estimations for different nodes. However, the overall imbalance loss can hinder the convergence of a group of neural networks (NNs) for generating balanced estimations. Additionally, numerous parameters need fine-tuning, such as the number of neurons, hidden layers, selection of activation functions, learning rate, and optimizer, which further complicate the learning process of NNs.

To solve the problem from the first step, following the initial estimations, our new algorithm, termed the Message Passing-based Node Weight Balancing Optimizer (MPO), takes over. MPO is designed to process the initial estimations and adjust them to yield more balanced outputs. The key advantages of MPO include its capability to utilize multi-processing techniques, O(MN) time and space complexity and data-driven mechanism. By processing samples in parallel, MPO significantly enhances the efficiency and speed of computation, making it particularly suitable for handling large datasets. Existing balancing methods, such as Balancing with Real Weights (BRW), do not account for the varying weights contributed by neighboring nodes\cite{rikos2014distributed,rikos2013distributed,priolo2013decentralized,gharesifard2009distributed,fang2019deep}. Additionally, traditional linear system-solving methods like CPLEX are not data-driven, meaning they produce consistent outputs only when the graph $G^{DF}$ remains unchanged\cite{cplex2009v12}.

%After the estimation adjustment by MPO, we employ a progressively enhanced regression model to iteratively learn the mapping between the input observations datasets and the adjusted estimations for variable nodes from MPO. For the regression models in this step, we offer a variety of models, such as LightGBM\cite{ke2017lightgbm}, XGBOOST\cite{Chen:2016:XST:2939672.2939785}, GBDT\cite{friedman2001greedy}, Random Forest(RF)\cite{ho1995random}, Decision Tree(DT), and Supervised Neural Networks(SNNs), LASSO, Ridge etc\cite{scikit-learn}. For ensemble learning models like LightGBM, we progressively enhance the regression model by increasing the number of decision trees or tree depth to address potential overfitting or underfitting issues. For neural network-based regression models, we load the trained model checkpoints from the first step to accelerate the learning process and apply L1, L2 regularization, and dropout strategies to mitigate overfitting and underfitting issues. Additionally, the default option is LightGBM, a powerful decision tree based machine learning algorithm. LightGBM was chosen for several reasons: its efficiency at handling large and complex data, the ability to prioritize important features through its built-in feature importance evaluation, and its robust regularization mechanisms including both L1 and L2 regularization. These features are crucial in many research areas where understanding the influence of specific data features on is paramount.

The trained regression model is then applied to perform inference on unseen data. This comprehensive framework, combining ssPINNs, MPO, and supervised learning model, offers a robust solution for researchers in the field of biology and beyond, aiming to provide deeper insights into the data characteristics. The integration of these methods not only enhances the accuracy and relevance of the data analysis but also accelerates the computational process, ensuring that results are both timely and scientifically meaningful. This methodological advancement thus represents a significant step forward in the application of computational techniques to complex real-world problems in various research fields, such as systems biology.

Algorithm \ref{algorithm-MPOCtrL} shows the framework of our method MPOCtrL, we apply the early stop strategy to avoid the trivial solution issue. The inputs include a DFG, a group of observation matrix datasets corresponding to the nodes in the DFG, the max epoch number $N_{max\_epoch}$ and imbalance loss threshold $\delta$ to stop the algorithm,  learning step $\beta$ and max epoch number $N_{max\_epoch}^{MPO}$ for MPO. Where $L_{im} = \sum_{V_{f_i} \in \{V_f\}} | \sum Parent Node(V_{f_i}) - \sum Child Node(V_{f_i}) | $ is the way to calculate the imbalance loss.

\begin{algorithm}
\SetAlgoLined
%\begin{algorithm}
%\ZZbaselinestretch{1.5}

\KwIn{$G^{DF},D=\{ X_k \}_{k=1}^K,\delta, N_{max\_epoch}, \beta, N_{max\_epoch}^{MPO}$}
\KwOut{$W$}
\While{$L_{im} > \delta \ or\ N_{current\_epoch} < N_{max\_epoch}$}{
    $W \leftarrow NNs(G^{DF}, \{ X_k \}_{k=1}^K)$\\
    $W \leftarrow (NNs+MPO)(G^{DF},W,\beta, N_{max\_epoch}^{MPO},\alpha)$ \# every 10 Epoches \\
    $L_{im} \gets L(G^{DF},W)$\\
    $N_{current\_epoch} \gets N_{current\_epoch} + 1$\\
}

\textbf{return} $W$

\caption{Message Passing Optimization-based Constrained Learning}
\label{algorithm-MPOCtrL}
\end{algorithm}

\subsubsection{Message Passing-based Node Weight Balancing Optimizer}
\label{mpo}

MPO is a message-passing-based algorithm for balancing the weights on factor graph nodes. It leverages \textbf{Belief Propagation} (BP), a fast and convergence-guaranteed algorithm to compute marginal probability on the nodes of a graph  ~\cite{yedidia2003understanding}. BP utilizes a sum-product strategy to calculate the marginal probability of nodes in time complexity $O(n)$ ~\cite{kschischang2001factor}. BP stands as a highly efficient and versatile inference technique within probabilistic graphical models, operating by passing messages between nodes and factors in a graphical model, thereby enabling the calculation of marginal and conditional probabilities. Its advantages encompass computational efficiency, exact inference capabilities in certain cases, flexibility across various graphical models, convergence guarantees, effective parallelization, and wide applicability in real-world domains such as computer vision, natural language processing, and communication networks. 

Adopting the idea of BP, we designed a novel formulation of MPO as detailed below to balance the weights on directed factor graph variables:

\begin{equation}
%\ZZbaselinestretch{1.5}
MSG_{V_{fa_i} \rightarrow V_{va_k}} = \vert \Sigma_{V_{va_m} \in NB(V_{fa_i})/V_{va_k}} (-1)^d \times MSG_{V_{va_m} \rightarrow V_{fa_i}} \vert
\label{equation-mpo-1}
\end{equation}

\begin{equation}
%\ZZbaselinestretch{1.5}
 MSG_{V_{va_i} \rightarrow V_{fa_k}} = \frac{(1-\beta) \times \Sigma_{V_{fa_m}\in NB(V_{va_i})/V_{fa_k}} MSG_{V_{fa_m} \rightarrow V_{va_i}}}{\vert NB(V_{va_i}) \vert -1} + \beta \times W_{V_{va_i}}
 \label{equation-mpo-2}
\end{equation}

\begin{equation}
%\ZZbaselinestretch{1.5}
W_{V_{va_i}} = \frac{\Sigma_{V_{fa_m}\in NB(V_{va_i})} \eta_m \times MSG_{V_{fa_m} \rightarrow V_{va_i}}}{\vert NB(V_{va_i}) \vert}
\label{equation-mpo-3}
\end{equation}

, where $d$ is -1 for variable $V_{va}$'s parent factors or 1 for $V_{va}$'s child factors. $\beta$ is the learning rate to update the variable. $\eta_m$ represents the imbalance level of factor $V_{fa_m}$, $NB$ denotes the neighbors, and $\|NB()\|$ means the number of neighbors.

\begin{algorithm}[ht]
\SetAlgoLined
%\begin{algorithm}
%\ZZbaselinestretch{1.5}

\KwIn{$G^{DF},W,\delta, N_{max\_epoch}^{MPO},\alpha$}
\KwOut{$W$}

\While{$L_{im} > \alpha \ or\ N_{current\_epoch} < N_{max\_epoch}^{MPO}$}{

    \For{i in 1,...,N}{
        Update $MSG_{fa_i \rightarrow va_j}, va_j\in NB(V_{fa_i})$ by (1)
    }

    \For{i in 1,...,M}{
        Update $MSG_{va_i \rightarrow fa_j}, f_j\in NB(V_{va_i})$ by (2)
    }

    \For{i in 1,...,M}{
        Update $W_{va_i}$ by (3)
    }

    $L_{im} \gets F_{imbalance\_loss}(G^{DF},W)$
    $N_{current epoch} \gets N_{current\_epoch} + 1$
}
\textbf{return}  $W$

\caption{Message Passing-based Node Weight Balancing Optimizer}
\label{algorithm-mpo}
\end{algorithm}

\textbf{Algorithm \ref{algorithm-mpo} } illustrate the main framework of MPO. The inputs of MPO include a directed factor graph $G^{DF}$, which has $N$ factors and $M$ variables, the initial weights on variables $W$. The max number of epochs $N_{max\_epoch}$, $\alpha$ is the threshold of the imbalance loss. The outputs are the balanced weights on nodes. 

\subsubsection{Illustration of the operations of MPO on a Directed Factor Graph}
\label{appendix_mpo}
MPO (Algorithm \ref{algorithm-mpo}) is an algorithm for optimizing and balancing the values of variables on a directed factor graph inspired by the message-passing operations in belief propagation. MPO algorithm balances messages over a DFG by iteratively operating the three steps: (1) passing messages from factor nodes to variable nodes, (2) passing messages from variable nodes to factor nodes, and (3) leveraging messages over the graph, as detailed below.

In the \textbf{first step}, we systematically traverse each factor within the $G^{DF}$. During this step, the key task is to update messages from the current factor to its neighboring variables. The update rules can be illustrated as follows: Consider a factor i connected to three neighboring variables—variable 1, variable 2, and variable 3, where variable 1 and variable 2 serve as the influx to factor i, while variable 3 is the outflux of factor i. To update the message from factor i to variable 1, we first focus on variable 1 and perform an operation known as masking, hiding variable 1. Then, we compute the value difference between variable 2 and variable 3, which forms the value of the message from factor i to variable 1 in the current update round. Following a similar rule, we proceed to update the message values from factor i to variable 2 and variable 3, and so forth. This iterative process continues to refine the messages exchanged between factors and variables, contributing to the optimization of the overall system.

In the \textbf{second} step of the MPO, we continue by iterating through each variable to update the messages sent from that variable to its neighboring factors. This step parallels the operation of the first step but inverts the direction of the message flow. While updating these messages, we apply a consistent procedure. When dealing with a variable's factor neighbors, we selectively mask one of the factors. Subsequently, we calculate the weighted mean of the messages originating from the remaining factor neighbors and directed toward the current variable. We sequentially update the messages based on this rule, ensuring a systematic flow from the current variable to its neighboring factors. This iterative process contributes significantly to message optimization and overall system refinement.

In the \textbf{third} step, we leverage the comprehensive set of obtained messages—ranging from all factors to their neighboring variables and from all variables to their neighboring factors. With this wealth of messaging data, the objective is to update the values assigned to each variable within the directed factor graph. This update process involves incorporating a scaling step $\beta$ applied to the initial value and $1-\beta$ applied to the new weight, $\beta$ ranges from (0,1). The new values are computed by averaging the messages originating from the neighboring factors to the current variable while considering their respective importance level $\eta$. This strategic adjustment of values based on message interactions is crucial for optimizing variable values to achieve balanced results that satisfy additive constraints. Specifically, this means that for each factor, the sum of its parent variables should be equal to the sum of its child variables.

\begin{figure*}[ht]

  \centering
  %\fbox{\rule[-.5cm]{0cm}{4cm} \rule[-.5cm]{4cm}{0cm}}
  \includegraphics[width=0.85\linewidth]{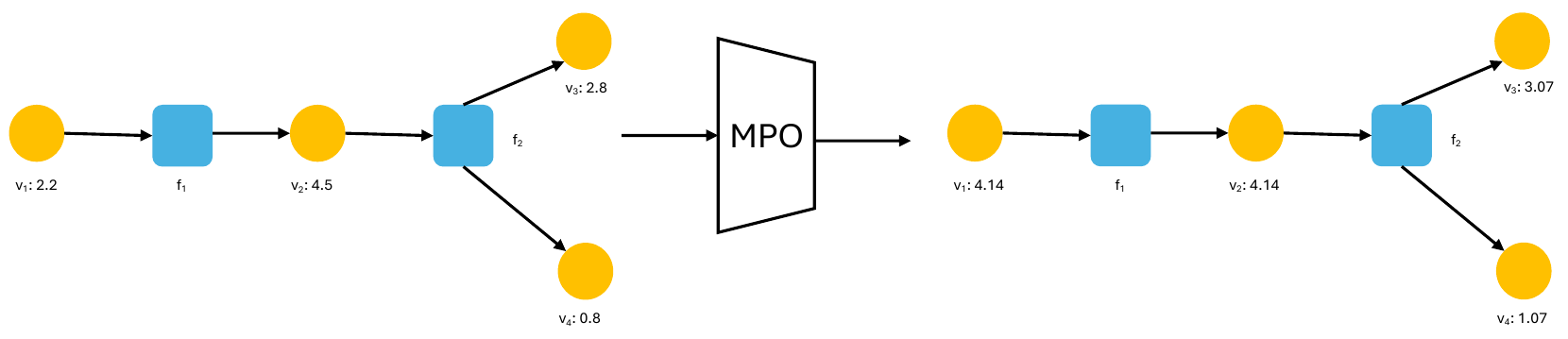}
  \caption{MPO Operations Example}\label{mpo_run_example}
\end{figure*}

Here, we show a \textbf{simple DFG} as illustrated in Fig \ref{mpo_run_example} to showcase how the MPO algorithm operates on a directed factor graph to balance the initial inputs:\\
\textbf{Init}:\\
(1) Variables: $v_1=2.2,v_2=4.5,v_3=2.8,v_4=0.8$\\
(2) Factors: $f_1:v_1=v_2,f_2:v_2=v_3+v_4$\\
Initial Messages are zeros.

\textbf{Epoch 1}:\\
(1)Update messages from variables to factors:\\
Messages from variables to factors:\\
$MSG_{v_1->f_1}=v_2=2.2$\\
$MSG_{v_2->f_1}=v_2=4.5$\\
$MSG_{v_2->f_2}=v_2=4.5$\\
$MSG_{v_3->f_2}=v_3=2.8$\\
$MSG_{v_4->f_2}=v_4=0.8$\\
(2)Update messages from factors to variables:\\
Messages from factors to variables: \\
$MSG_{f_1->v_1}=MSG_{v_2->f_1}=4.5$\\
$MSG_{f_1->v_2}=MSG_{v_1->f_1}=4.5$\\
$MSG_{f_2->v_2}=MSG_{v_3->f_2}+MSG_{v_4->f2}=2.8+0.8=3.6$\\
$MSG_{f_2->v_3}=MSG_{v_2->f_2}-MSG_{v_4->f2}=4.5-0.8=3.7$\\
$MSG_{f_2->v_4}=MSG_{v_2->f_2}-MSG_{v_3->f_2}=4.5-2.8=1.7$\\
(3)Update variables, $\beta=0.5,\eta=1$:\\
$v_1=(1-\beta) \times v_1 + \beta \times \frac{\eta \times MSG_{f_1->v_1}}{1}=\frac{2.2+4.5}{2}=3.35$\\
$v_2=(1-\beta) \times v_2 + \beta \times \frac{\eta \times MSG_{f_1->v_2}+MSG_{f_2->v_2}}{2}=\frac{4.5+3.6}{2}=4.05$\\
$v_3=(1-\beta) \times v_3 + \beta \times \frac{MSG_{f_2->v_3}}{1}=\frac{2.8+3.7}{2}=3.25$\\
$v_4=(1-\beta) \times v_4 + \beta \times \frac{MSG_{f_2->v_4}=1.7}{1}=\frac{0.8+1.7}{2}=1.25$\\

\textbf{Epoch 2}:\\
(1)Update messages from factors to variables:\\
Messages from factors to variables: \\
$MSG_{f_1->v_1}=MSG_{v_2->f_1}=4.05$\\
$MSG_{f_1->v_2}=MSG_{v_1->f_1}=4.05$\\
$MSG_{f_2->v_2}=MSG_{v_3->f_2}+MSG_{v_4->f2}=3.25+1.25=4.5$\\
$MSG_{f_2->v_3}=MSG_{v_2->f_2}-MSG_{v_4->f2}=4.05-1.25=2.8$\\
$MSG_{f_2->v_4}=MSG_{v_2->f_2}-MSG_{v_3->f_2}=4.05-3.25=0.8$\\
(2)Update messages from variables to factors:\\
Messages from variables to factors:\\
$MSG_{v_1->f_1}=MSG_{f_1->v_1}=v_2=4.05$\\
$MSG_{v_2->f_1}=MSG_{f_2->v_2}=v_2=4.05$\\
$MSG_{v_2->f_2}=MSG_{f_1->v_2}=v_2=4.04$\\
$MSG_{v_3->f_2}=MSG_{f_2->v_3}=v_3=3.25$\\
$MSG_{v_4->f_2}=MSG_{f_2->v_4}=v_4=1.25$\\
(3)Update variables, $\beta=0.5,\eta=1$:\\
$v_1=(1-\beta) \times v_1 + \beta \times \frac{\eta \times MSG_{f_1->v_1}}{1}=\frac{3.35+4.05}{2}=3.7$\\
$v_2=(1-\beta) \times v_2 + \beta \times \frac{\eta \times MSG_{f_1->v_2}+MSG_{f_2->v_2}}{2}=\frac{4.05+4.5}{2}=4.275$\\
$v_3=(1-\beta) \times v_3 + \beta \times \frac{MSG_{f_2->v_3}}{1}=\frac{3.25+2.8}{2}=3.025$\\
$v_4=(1-\beta) \times v_4 + \beta \times \frac{MSG_{f_2->v_4}=1.7}{1}=\frac{1.25+0.8}{2}=1.025$\\

\textbf{Epoch 3}:\\
...\\

Finally, $v_1\approx 4.14,v_2\approx4.14,v_3\approx3.07,v4\approx3.07$, which are balanced on the DFG shown in Fig \ref{mpo_run_example}.

\begin{figure*}[ht]
  \centering
  %\fbox{\rule[-.5cm]{0cm}{4cm} \rule[-.5cm]{4cm}{0cm}}
  \includegraphics[width=\linewidth]{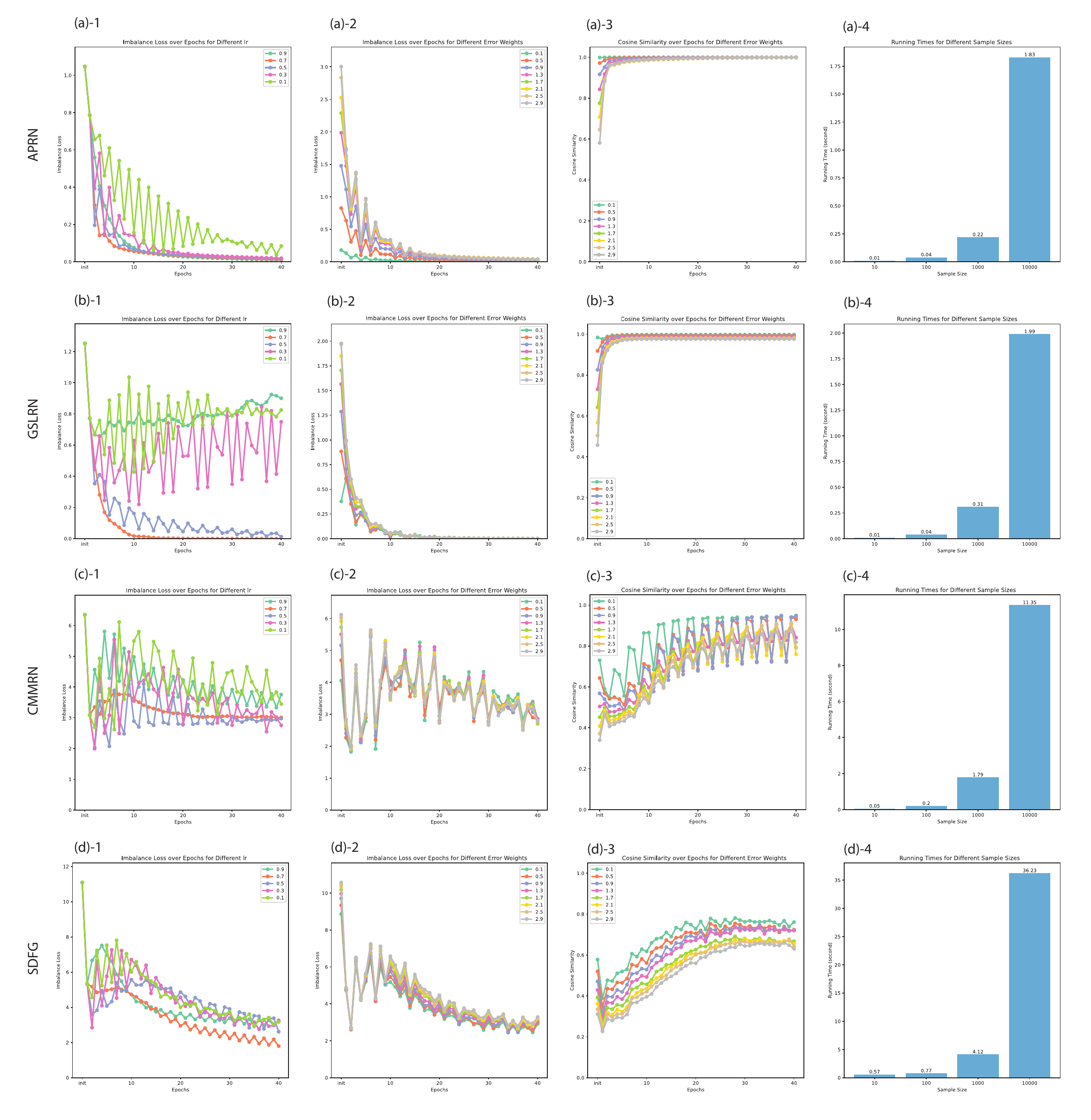}
  \caption{Experiments about Effectiveness, Robustness, Running Time of MPO on four Directed Factor Graphs}\label{mpo_details}
\end{figure*}

\begin{figure}
  \centering
  %\fbox{\rule[-.5cm]{0cm}{4cm} \rule[-.5cm]{4cm}{0cm}}
  \includegraphics[width=1.0\linewidth]{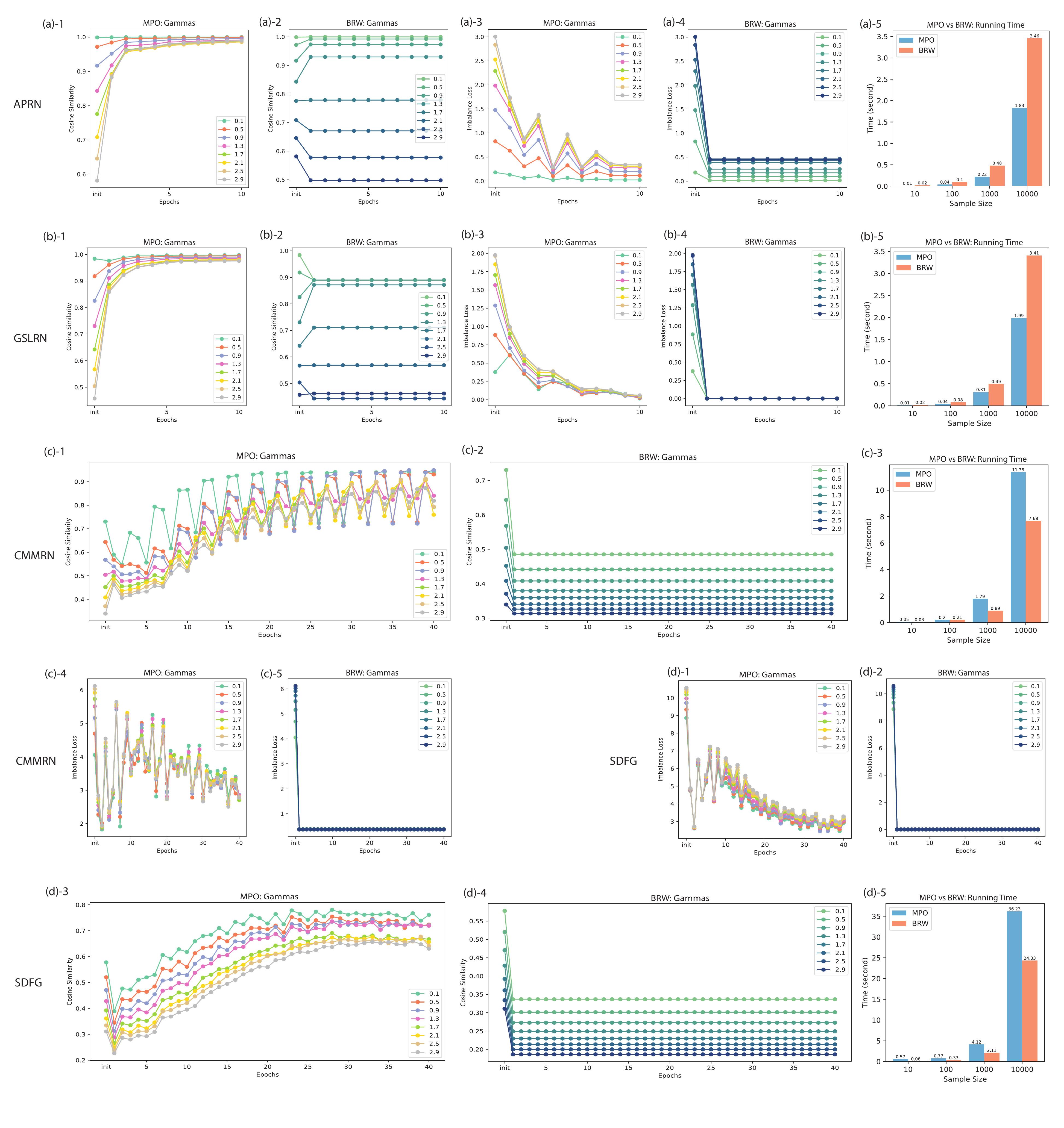}
  \caption{Comparisons between MPO and BRW on four directed factor graphs for different error level gammas and running time}\label{mpo_vs_brw_appendix}
\end{figure}

\subsection{Mathematical discussion of MPO}
\textit{Scalability}\\
The MPO algorithm prioritizes high efficiency and optimized runtime performance. It systematically iterates through all factors and their respective neighboring nodes, then through all nodes and their neighboring factors. Core operations such as addition, subtraction, and averaging maintain a constant time complexity of $O(1)$.

Considering a Directed Factor Graph (DFG) with $N$ factor and $M$ variable nodes, the algorithm's worst-case scenario involves each node being connected to every factor. Two primary computations are crucial: messages flowing from factors to nodes and vice versa. To enhance efficiency, these messages are computed once and stored, leveraging caching to avoid redundant recalculations. This approach ensures consistency in message usage, significantly reducing computational overhead and enhancing overall efficiency. Consequently, the MPO algorithm's time complexity is determined to be $O(MN)$, reflecting its suitability for practical applications.

Other than time complexity, it is also imperative to evaluate space complexity in the MPO algorithm. MPO retains all message information during execution and covers messages from each factor to its neighboring nodes and vice versa. As a result, the space complexity of the MPO algorithm is $O(MN)$.

\textit{Global optimization over the input DFG}\\
The MPO procedure updates a node's weights by integrating messages from all neighboring factors directed towards that node. These messages contain information about the weights of all nodes in their direction. Consequently, when updating a node's weights, MPO doesn't rely solely on local information but integrates data from all nodes and factors, excluding the to-be-updated node. This global perspective allows MPO to consider the entire network, enhancing its effectiveness compared to a narrow, node-centric approach.

\textit{Convergence and stop criterion}\\
In a balanced graph, the total weights of input nodes equal those of output nodes. Leveraging the global perspective in the MPO algorithm, messages from factors to nodes carry crucial adjustments and updated weights of neighboring nodes to attain equilibrium within the current factor. Unlike localized updates, MPO takes a global approach that advances the entire graph towards balance in each iteration. This equilibrium is achieved by minimizing differences between the weights of incoming and outgoing nodes. Iterations will be terminated when this imbalance is close to zero.

\subsection{Experimental Details}
\subsubsection{Networks analyzed in the synthetic data based experiments} 
We tested our methods on four real-world biological reaction networks, each containing metabolites (factors) and reactions (variables) that are crucial to cancer treatment \cite{zhang2023fluxestimator}. Additionally, we evaluated our approach on a highly complex synthetic directed factor graph. 

The details of each network are as follows:\\
(1) Antigen Presentation Reaction Network (APRN) includes 10 metabolites, 11 reactions, please see fig \ref{aprn}. APRN is an important process in the immune system where specialized cells capture foreign substances, called antigens, and present them to T cells through major histocompatibility complex (MHC) molecules\cite{blum2013pathways}. (2) Glutamine Sub cellular Localization Reaction Network (GSLRN), contains 6 metabolites and 10 reactions see fig \ref{gslrn}. (3) Central Metabolic Map Reaction Network (CMMRN) was encapsulated in scFEA\cite{alghamdi2021graph} comprises 66 metabolites, 159 reactions, and 3 cycles see fig \ref{cmmrn}. (4) Glutamine Glucose Subcellular Localization Reaction Network (GGSLRN), contains 23 metabolites and 42 reactions see fig \ref{ggslrn}, This network was exclusively applied to real-world data due to its relevance and applicability to the specific disease being studied.  (5) Synthetic Directed Factor Graph (SDFG), a more complicated synthetic network. This synthetic network has 204 factors, 453 variables, and 18 cycles please see \ref{sdfg}. For more information of these networks, please refer to the APPENDIX and Supplementary Material. For the real-world biological reaction networks, the count and name of genes (features) in the reactions (variables) are also known. Then we converted metabolites into factors and reactions into variables to match the $G^{DF}$ format for the experiments. Each factor represents a linear relationship between its parent and child variables. Each variable contains observation data in matrix format, where rows represent samples and columns represent features, along with the estimations for the samples we need to estimate. Here, in the biological cases, the samples could be cells, features could be genes involved in the reaction, and the estimations could be the fluxes for the cells in the reaction.

\subsubsection{Generation of Synthetic Observation Data.} 
For a given network of $n$ metabolites, i.e., factor nodes $x_i,i=1...,n$, and $K$ reactions, i.e., variable nodes $f_k,k=1,...,K$, we simulate $D_{j,.}, j=1,...,m$ as non-time course observations, each associated with the $n$ factor nodes. Note that $D_{j,.}$ may not be necessarily direct observations of the metabolites $x_i,i=1,...,n$.  We first simulate a large matrix denoted as $Y^{(m\times K)}$, where each row $Y_{j,.}$ is the underlying truth of flux for all the reactions $f_1,...,f_K$, for sample $j$. To simulate each $Y_{j,.}$, we randomly generate a set of positive numbers and input them into the MPO algorithm to obtain a set of balanced values on a directed factor graph. In other words, for each factor node, its input flux and output flux are equal, and hence balanced. We then simulate the $D_{j,.}$ based on the underlying $Y_{j,.}$, in a backward fashion. Entries in each $D_{j,.}$ and $Y_{j,.}$ are linked by non-linear functions. Two non-linear formulas (NLFs) were used: (1) NLF1, $Y_{j,l} = a \times (\sum_{i\in Z_l} D_{j,z})^2 + b \times (\sum_{z\in Z_l} D_{j,z})$; and (2) NLF2, 
$Y_{j,l} = \exp\{a \times (\sum_{i\in Z_l} D_{j,z})^2 + b \times (\sum_{i\in Z_l} D_{j,z})\}$. Here $Y_{j,l}$ denotes the true flux for the $l$-th reaction in the $j$-th sample; $D_{j,z}$ denotes the $z$-th attribute value in the $j$-th sample; $Z_l$ indicates all those node attributes in $D_{j,.}$ that is associated with the $l$-th reaction; $a,b$ are randomly generated constants ranging from (0,10). With simulated $Y_{j,.}$ and the link functions NLF1 and NLF2, we could obtain many solutions of $D_{j,.}$. A random solution of $D_{j,.}$ is picked, and 20\% of its values are set to zero to simulate data sparsity, then the vector is normalized to ensure they sum up to 1. Finally, the directed factor graph, together with the simulated $D$ are passed onto each method, to solve for $Y$, which is then compared with the truth to evaluate the accuracy of the methods.

\subsection{Experiments to evaluate the Robustness and Effectiveness of MPO}
\textbf{Robustness} is exemplified through its performance on synthetic data, even when subjected to varying levels of error. This evaluation was crucial to demonstrate MPO's resilience and reliability in practical scenarios where data imperfections are common. By introducing errors into the synthetic datasets at different magnitudes, we aimed to mimic real-world challenges and test MPO's adaptability. The outcomes were compelling, with MPO consistently achieving results that exhibited high cosine similarity with the target configurations, regardless of the error level. This consistent performance underlines MPO's robustness against data inaccuracies, highlighting its potential for widespread application. The ability of MPO to maintain high fidelity in optimization outcomes, despite data noise, positions it as a highly reliable method for graph-based optimization tasks. Such robustness is indicative of MPO's capability to deliver accurate and dependable results, making it an invaluable tool in scenarios where data may not be pristine. Overall, MPO's resilience to errors further strengthens its case as a robust and versatile optimizer for factor graphs, ensuring reliability and accuracy even in the face of data imperfections.

Fig \ref{mpo_details} (a-1, b-1, c-1, d-1) illustrates the sensitivity of the MPO algorithm to different learning rates $\beta={0.1,0.3,0.5,0.7,0.9}$ across four different directed factor graphs, ranging from simple to complex. The learning rate $\beta$ determines the ratio at which information from neighbors is accepted to update the current variable. In this analysis, all initial variables are randomly generated to ensure a sufficiently high initial imbalance loss. The figures consistently demonstrate that the imbalance loss decreases steadily, highlighting the robustness of the MPO algorithm across various learning rates and graph complexities.

\textbf{Effectiveness} The Message Passing-based Optimizer (MPO) stands out in balancing node weights on factor graphs through its efficient use of message passing techniques, effectively transitioning from initial imbalances to a uniform distribution. The core of MPO's approach lies in iteratively refining node weights by integrating information from neighboring nodes, fostering a balance that is informed on a global scale. This iterative process is sustained until the disparity in node weights is significantly reduced, showcasing MPO's ability to attain optimal balance swiftly and with minimal resource consumption. Notably, the transformation facilitated by MPO, from a starkly imbalanced initial state to a uniform distribution, is vividly illustrated in Figure 1, emphasizing the method's effectiveness in large-scale graphs. Moreover, MPO's operational excellence is further underscored by its computational efficiency, with both time and space complexities pegged at \(O(MN)\), where \(M\) represents the number of factors and \(N\) the number of nodes. This scalability and systematic approach position MPO as a pivotal advancement in graph-based optimization, adept at addressing weight imbalance through a principled mechanism. 

Fig \ref{mpo_details} (a-2,a-3,a-4), (b-2,b-3,b-4),(c-2,c-3,c-4),(d-2,d-3,d-4) show MPO's performance across various metrics: cosine similarities between targets $W$ and recovered weights, imbalance loss, and running time.  For the recovery experiments, we followed $Error= \gamma \sum_{i=1}^{3} a_i v_i + \epsilon$ and $W=\frac{W}{norm(W)}+Error$ to simulate added error inputs to MPO, ensuring that the inputs had controlled imbalance loss and cosine similarity that the targets deviated from its original geometrical space. For the target $W$, we randomly generate a set of positive numbers and input them into the MPO algorithm to obtain balanced initial values on a directed factor graph. This process ensures that the initial values are adjusted appropriately within the graph structure to meet the balancing criteria. Where $v_i$ is the orthogonal vector to $W$, $\gamma = \{0.1,0.5,0.9,1.3,1.7,2.1,2.5,2.9\}$ represents different error weight levels to the orthogonal vectors, $a_i$ is a random float number ranges from (0,1), $\epsilon$ is a constant number 0.1. This deviation was measured by the cosine similarity between the targets and the methods' outputs. These figures demonstrate that, over epochs, the imbalance loss steadily decreases while the cosine similarity steadily increases. This trend confirms the correctness and effectiveness of the MPO algorithm across four different directed factor graphs, ranging from simple to complex. The consistent improvement in both imbalance loss and cosine similarity highlights the algorithm's robustness and ability to optimize variable values efficiently.

Fig \ref{mpo_vs_brw_appendix} compare MPO with an existing weight balancing method, BRW, across various metrics: cosine similarities between targets $W$ and recovered weights, imbalance loss, and running time. Here, we assess given an initial value to MPO, how well it could recover the ground truth. To test MPO's robustness to noise, we followed $Error= \gamma \sum_{i=1}^{3} a_i v_i + \epsilon$ and $W=\frac{W}{norm(W)}+Error$ to simulate added error inputs to MPO and BRW, ensuring that the inputs had controlled imbalance loss and cosine similarity that the targets deviated from its original geometrical space. For the target $W$, we randomly generate a set of positive numbers and input them into the MPO algorithm to obtain balanced initial values on a directed factor graph. This process ensures that the initial values are adjusted appropriately within the graph structure to meet the balancing criteria. Where $v_i$ is the orthogonal vector to $W$, $\gamma = \{0.1,0.5,0.9,1.3,1.7,2.1,2.5,2.9\}$ represents different error weight levels to the orthogonal vectors, $a_i$ is a random float number ranges from (0,1), $\epsilon$ is a constant number 0.1. This deviation was measured by the cosine similarity between the targets and the methods' outputs. Our method, MPO, outperformed BRW in terms of recovering cosine similarities in all the datasets, ranging from simple to complex. BRW simply calculates the sum of weights for parent variables, then assigns these weights to the child variables by dividing the sum by the number of child variables and adding the original weights of the child variables. This approach does not account for the varying importance of variables and factors within the $G^{DF}$, whereas MPO considers these differences by involving neighbors' imbalance levels. We did not compare our MPO method with some linear system solving methods like CPLEX because these methods are not data-driven. When the linear systems are fixed, they produce identical solutions for different samples, lacking the adaptability to variations in the data.

\begin{figure}
  \centering
  %\fbox{\rule[-.5cm]{0cm}{4cm} \rule[-.5cm]{4cm}{0cm}}
  \includegraphics[width=1.0\linewidth]{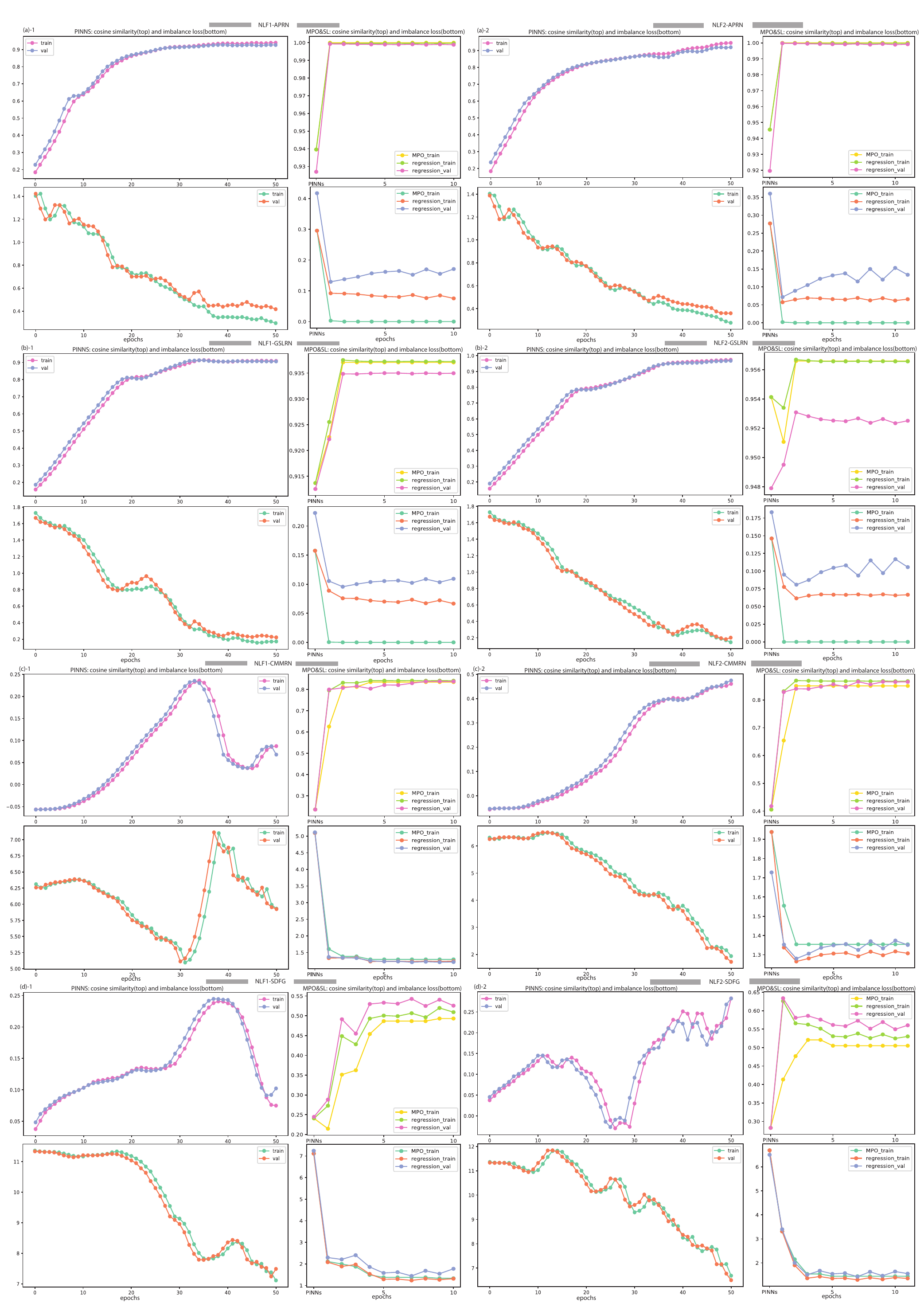}
  \caption{Experiments of our method on Full Synthetic Observation Data}\label{MPOCtrL_appendix}
\end{figure}

\subsection{Details of experiments to Benchmark the robustness and accuracy of MPOCtrL}
The first step, ssPINNs, involves a group of neural networks. Each neural network consists of a single hidden layer with 16 neurons, uses the tanhshrink activation function and Adam optimizer, learning rate is 0.08. The output layer of each network has one neuron. The outputs from these neural networks are then collected to calculate the loss. Fig \ref{MPOCtrL_appendix} demonstrates the robustness and accuracy of ssPINNs on both the training and validation datasets. Left of fig \ref{MPOCtrL_appendix} (a1, a2; b1, b2; c1, c2; d1, d2) shows the mean of sample-wise cosine similarities between targets and estimations(top), and imbalance loss(bottom) over epochs for synthetic datasets NLF1-APRN, NLF2-APRN, NLF1-GSLRN, NLF2-GSLRN, NLF1-CMMRN, NLF2-CMMRN, NLF1-SDFG, NLF2-SDFG, respectively. When the network is simple, such as APRN and GSLRN, the first step, ssPINNs cloud steadily generate higher cosine similarities and lower imbalance losses over epochs. However, for more complex networks like CMMRN and SDFG, which contain more factors, variables, and cycles, ssPINNs struggle. This issue arises because ssPINNs rely on a single total loss combined from all NNs, which is insufficient to effectively guide the learning process in complex networks. Despite this, Right of figures \ref{MPOCtrL_appendix} (a1, a2; b1, b2; c1, c2; d1, d2) demonstrate that the second step, MPO, and the third step, supervised learning(SL), can still achieve convergence based on the initial estimations provided by ssPINNs. To avoid zero solutions and over-fitting, we applied an early stopping strategy. These figures illustrate how well our approach performs across different complexities of data, ensuring the reliability and correctness of the model's estimations. 

\subsection{Coherency loss $L_\Phi$ and Approximation of Ordinary Dynamic Models for Different Types of Systems and Data.}

We classify data-driven approximation of dynamic models into sub-tasks based on the dynamic property of systems and observed data types. We first classify dynamic systems into systems of equilibrium steady states (ESS) or non-equilibrium steady states (NESS). An ESS system is defined by having all variables trackable and unchanged under a steady state. For NESS system, we consider all the variables \( x_i \) to be capped by an upper bound. Also, we will consider the context-specific status of the system as under steady or quasi-steady state (SS) and dynamic state (DS). Upon the observed data, we consider if the variables are directly observed or unobserved and if the data is of time-course or non-time-course. \textbf{Table 2} summarizes the different sub-tasks.

\begin{table}[ht]
\centering
\caption{System Classification}\label{table:classification}
\begin{tabular}{|p{3cm}|c|c|c|c|c|}
\hline
\rowcolor{gray!20}
\textbf{System Classification} & \multicolumn{2}{c|}{\textbf{Steady State (SS)}} & \multicolumn{2}{c|}{\textbf{Dynamic State (DS)}} \\ \cline{2-5} 
 & \textbf{Time course} & \textbf{Non-time course} & \textbf{Time course} & \textbf{Non-time course} \\ \hline
\multicolumn{1}{|p{3cm}|}{\textbf{Equilibrium Steady State (ESS)}} & \multicolumn{4}{c|}{E.g., Mass carrying flux system} \\ \cline{2-5} 
 & Observed & Unobserved & Observed & Unobserved \\
 & Time course & Non-time course & Time course & Non-time course \\
 \hline
\multicolumn{1}{|p{3cm}|}{\textbf{Non-Equilibrium Steady State (NESS)}} & \multicolumn{4}{c|}{E.g., Signal amplification system} \\ \cline{2-5} 
 & Observed & Unobserved & Observed & Unobserved \\
 & Time course & Non-time course & Time course & Non-time course \\
 \hline
\end{tabular}
\end{table}

\textbf{Sub-task 1: ESS system under SS.}
Most mass-carrying networks, such as metabolic pathways or traffic flow, are ESS systems. The formula \textcircled{2} provides one basic loss term $L_\Phi$ for the ESS system under SS condition. However, it only considers the Law of conservation of mass as the constraint. More constraints can be brought in by considering dynamic properties of \(F\), such as the non-linear dependency between reaction rate and substrates, enzymes, and co-factors suggested by the Michaelis-Menten equation. The following loss term could be specifically introduced:
\begin{equation}
L_{ES-ODE} = \sum_{j=1}^N \sum_{m=1}^M \sum_{i=1}^{i_m} \left( \frac{\partial F_m}{\partial g_i^m} (D_j^m\mid \Theta) - \frac{\partial R_m}{\partial g_i^m} (X_j^m ) \right)^2,
F_m (D_j^m\mid \Theta) = f_{nn}^m (D_j^m \mid \theta_m),
\end{equation}

where \( \frac{\partial R_m}{\partial g_i^m} (X_j^m ) \)
denotes the theoretic dependency between the real dynamic model of the reaction \( R_m \) and its molecular feature \( D_j^m \) that can be reflected by the function of \( \Theta \) and \( D_j^m \).

\textbf{Sub-task 2: \textbf{ESS System Under DS}}. Four scenarios are considered here. 

\textbf{Scenario 1:} if both time course data and the variables are observed, denote observed data as \( D_t = \{X_t, D_t\} \), where \( X_t \), \( X_t \), and \( t = \{t_1, \ldots, t_N\} \) represent observed variables, other features, and time points, respectively. We continue to use the notation \( FG(C,R,E_{X \to R}, E_{R \to X}) \) for a directed factor graph-based representation of the system as described ealier. For each molecule \( X_k \), denote \( X_{k,t_j} \) as its observed value at time \( t_j \). For other molecular features \( D_t \), denote \( D_{t_j}^m = \{D_{1,t_j}^m, \ldots, D_{t_j}^m\} \) as the other molecular features involved in the reaction \( R_m \) observed at \( t_j \). The reaction rate of \( R_m \) is modeled as \( F_{m,t_j} = f_{nn}^m (X_{t_j}^{m_{IN}}, D_{t_j}^m \mid \theta_m) \) as a multi-layer neural network with the input \( D_{t_j}^m \) and \( X_{t_j}^{m_{IN}} \), where \( X_{t_j}^{m_{IN}} \) denotes the molecules serves as the input of \( R_m \), and \( \theta_m \) is the parameter of the neural network. \( \theta_m \) and cell-wise flux \( F_{m,j} \) could be solved by the loss:

\begin{equation}
\small
L_{ESS-DS} = \sum_{j=1}^{N-1} \sum_{k=1}^K H_k \left(\frac{dX_k}{dt}\bigg|_{t=t_j}, X_{k,t_{j+1}} - X_{k,t_j} \right)
\label{eq:loss}
\end{equation}
\begin{equation}
\frac{dC_k}{dt}\bigg|_{t=t_j} \overset{\Delta}{=} \sum_{m' \in R_{\text{out}}^{X_k}} F_{m',j} (X_{t_j}^{m_{IN}'}, D_{t_j}^{m'}) - \sum_{m \in R_{\text{in}}^{X_k}} F_{m,j} (X_{t_j}^{m_{IN}}, D_{t_j}^m) 
\label{eq:loss2}
\end{equation}
,where \( H_k \) measures the difference between the predicted changing rate of \( X_k \) at time \( t_j \), \( \frac{dX_k}{dt}\bigg|_{t=t_j} \) and the observed change of \( X_k \) between \( t_{j+1} \) and \( t_j \). If the observational error of \( X_k \) is Gaussian, \( H_k = (X_{k,t_{j+1}} - X_{k,t_j} - \frac{dX_k}{dt}\bigg|_{t=t_j} \cdot (t_{j+1} - t_j))^2 \).

\textbf{Scenario 2:} If the products of the reactions are observed but there is not time course information, we will align the samples by predicting their SS and pseudo-time. A pseudo-time \( t_{i,j} \) will be assigned to sample \( j \) in the \( i \)th sample group by the distance between the sample’s current state and its SS. 

\textbf{Scenario 3:} If time course information is available but the products of the reactions are not observed, we will consider \( X_{k,t_j} \) as latent variables and iteratively update \( X_{k,t_j} \) and \( F_{m,j} \). 

\textbf{Scenario 4:} if both time course information and the products of the reactions are not observed, we will integrate the solution for scenarios 2 and 3 by iteratively estimating \( F_{m,j} \), \( X_{k,t_j} \), and \( t_{i,j} \).

Noted, MPOCtrL demonstrated that the BP-based MPO can effectively handle linear constraints such as flux balance under a quasi-steady state. By using this approach, the constraints of dynamic properties and the principle of parsimony could be handled by supervised learning in Step 3 of the MPO-SL framework. It is noteworthy that the MPO (Step 2) forms an estimation of Steady State (SS) for Equilibrium Steady State (ESS) systems when the input is not under SS state. For a given numerical solution of the flux rates in a system, $\mathcal{F}(D, \Theta)$, which is not under the SS state, MPO could approximate a numerical solution of its flux rate under SS, $F'$, by minimizing $L_\Phi(\mathcal{F}(D, \Theta))$. Thus, MPO could be applied to predict the SS for each sample and further estimate the distance of each sample to their SS for the ESS system under Dynamic States (DS).

\textbf{Sub-task 3: NESS System Under SS}. For a NESS system, under SS, time-course information is unnecessary. However, the product must be observed to ensure identifiability. The following supervised learning loss will be utilized to link changes in molecular features and the reaction rate:
\begin{equation}
L_{\text{NESS}} = \sum_{j=1}^N \sum_{k=1}^K (X_{k,j} - \sum_{m \in R_{\text{in}}^{X_k}} F_{m,j})^2 \label{eq:NESS}
\end{equation}

\textbf{Sub-task 4: NESS System Under DS}.
Observation of the products of each reaction is necessary to ensure identifiability. If time course data is available, \(F_{m,j}\) could be estimated by minimizing the loss term \eqref{eq:loss}. If time course data is not available, then if there is an approach to reliably estimate sample-wise pseudo-time, such as the trajectory and pseudo-time inference for scRNA-seq data of a development system, we will use sample-wise pseudo-time and apply the loss term \eqref{eq:loss} to estimate \(F_{m,j}\).

\begin{figure}
  \centering
  %\fbox{\rule[-.5cm]{0cm}{4cm} \rule[-.5cm]{4cm}{0cm}}
  \includegraphics[width=0.7\linewidth]{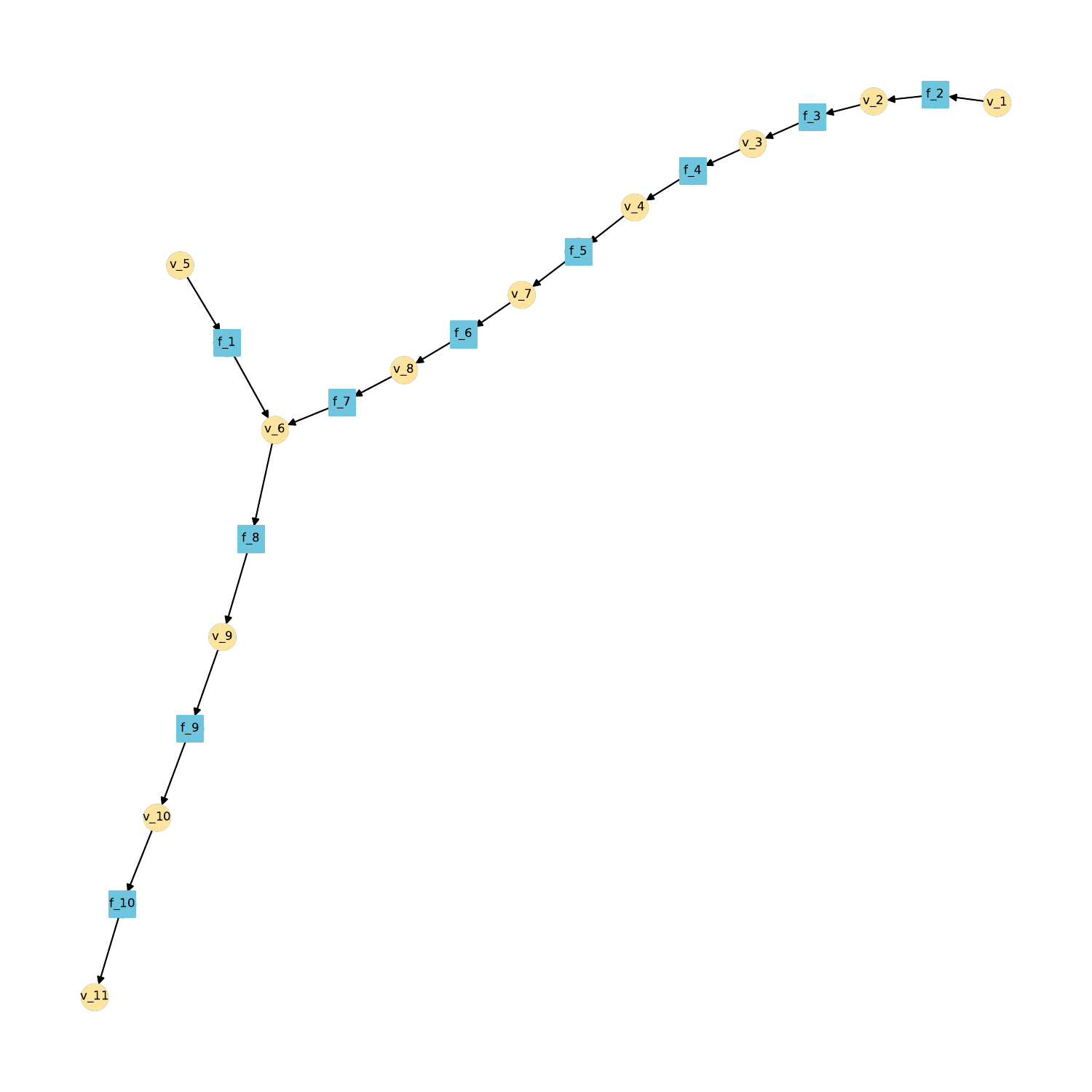}
  \caption{Antigen Presentation Reaction Network (APRN)}\label{aprn}
\end{figure}

\begin{figure}
  \centering
  %\fbox{\rule[-.5cm]{0cm}{4cm} \rule[-.5cm]{4cm}{0cm}}
  \includegraphics[width=0.7\linewidth]{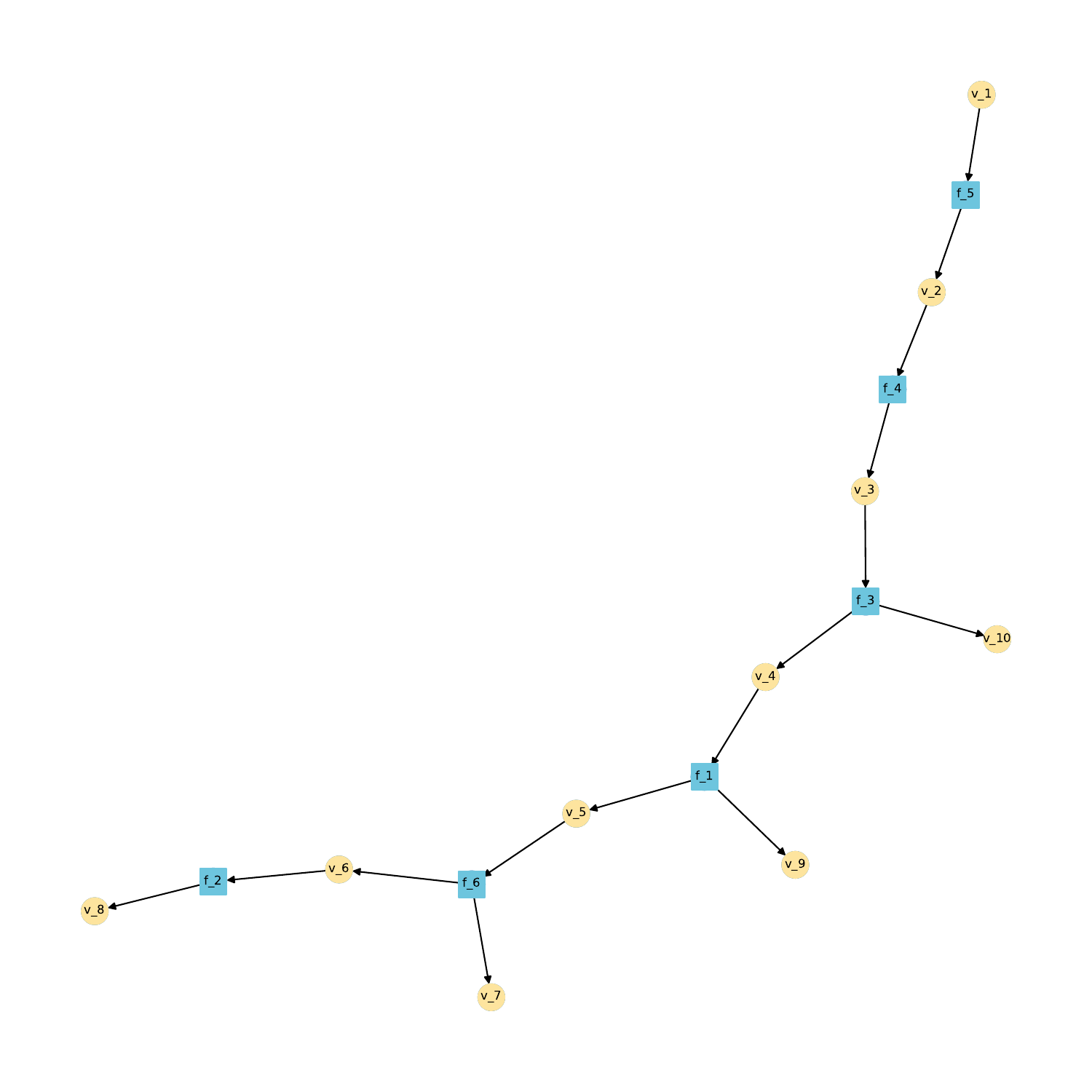}
  \caption{Glutamine Sub-cellular Localization Reaction Network (GSLRN)}\label{gslrn}
\end{figure}

\begin{figure}
  \centering
  %\fbox{\rule[-.5cm]{0cm}{4cm} \rule[-.5cm]{4cm}{0cm}}
  \includegraphics[width=0.7\linewidth]{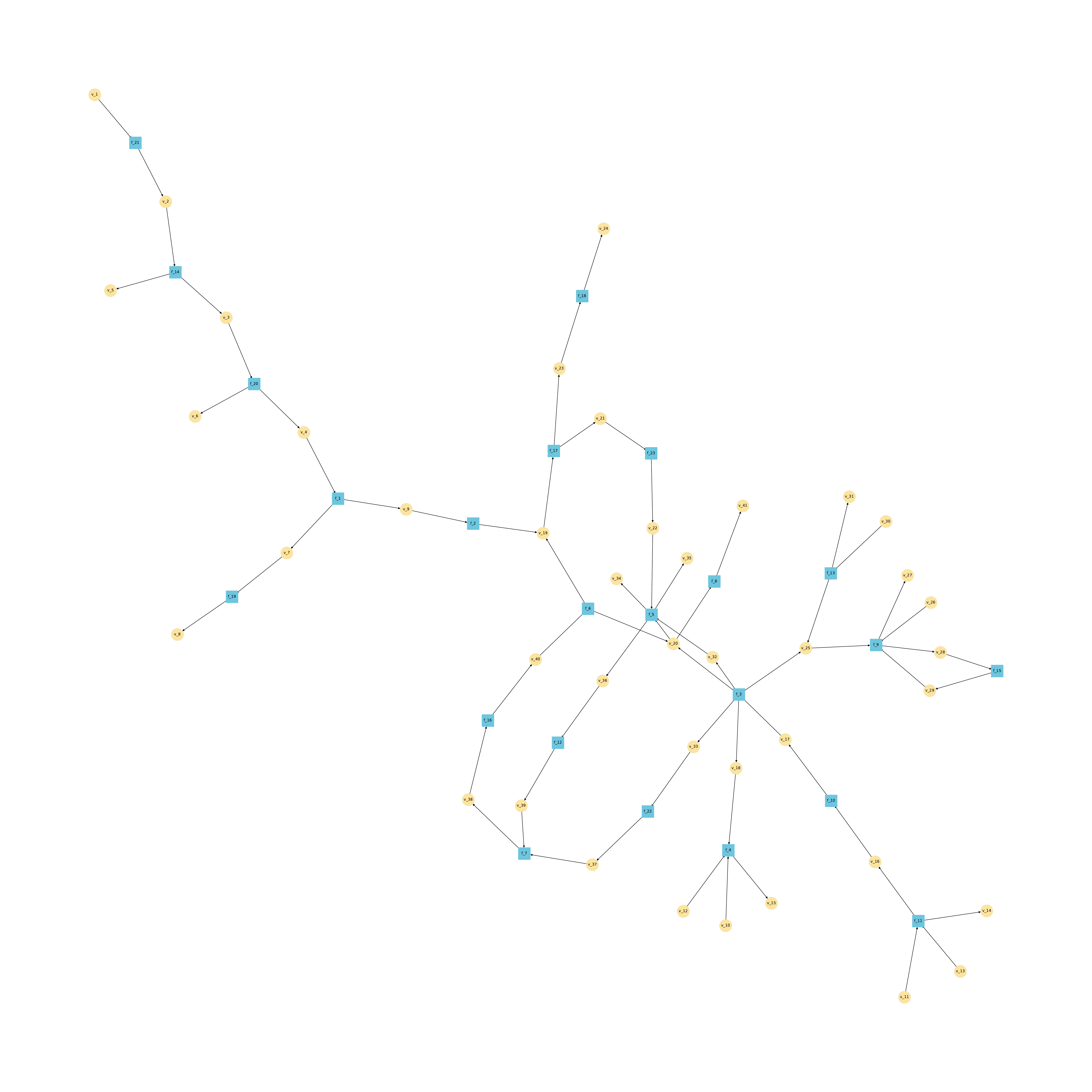}
  \caption{Glutamine Glucose Sub-cellular Localization Reaction Network}\label{ggslrn}
\end{figure}

\begin{figure}
  \centering
  %\fbox{\rule[-.5cm]{0cm}{4cm} \rule[-.5cm]{4cm}{0cm}}
  \includegraphics[width=1.0\linewidth]{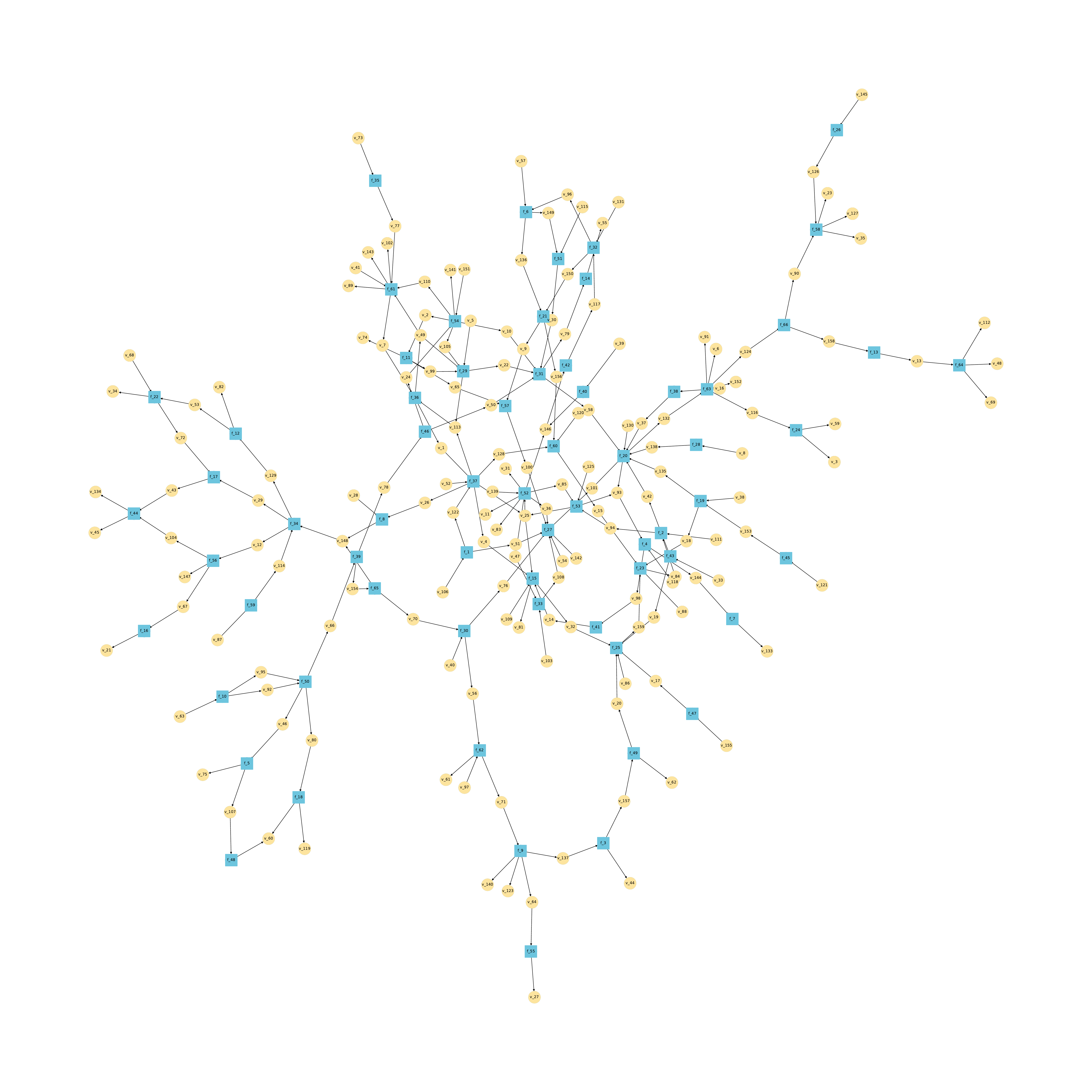}
  \caption{Central Metabolic Map Reaction Network (CMMRN)}\label{cmmrn}
\end{figure}

\begin{figure}
  \centering
  %\fbox{\rule[-.5cm]{0cm}{4cm} \rule[-.5cm]{4cm}{0cm}}
  \includegraphics[width=1.0\linewidth]{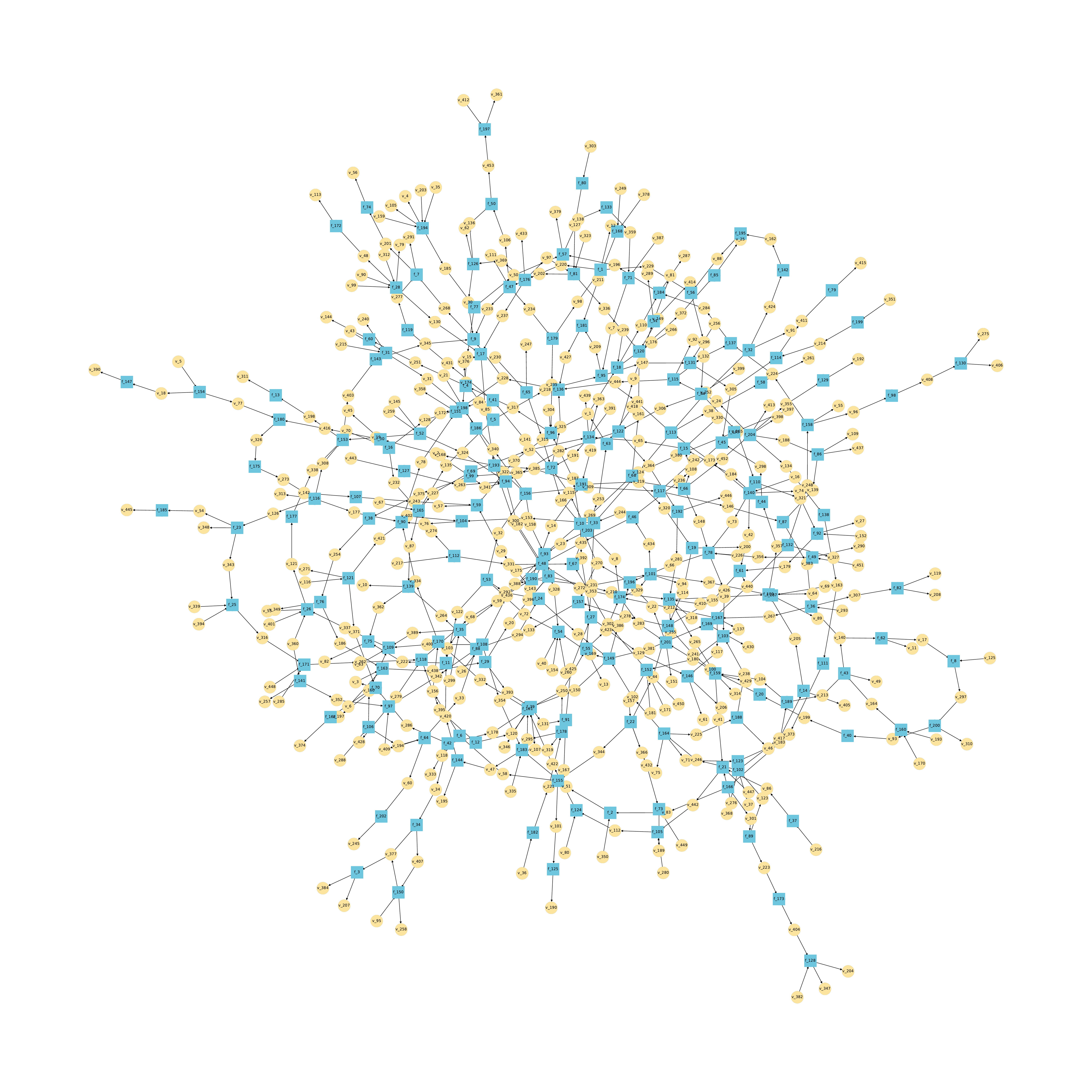}
  \caption{Synthetic Directed Factor Graph (SDFG)}\label{sdfg}
\end{figure}

\end{document}